\pdfoutput=1
\documentclass[11pt]{article}

\usepackage{acl}

\usepackage{times}
\usepackage{latexsym}
\usepackage{booktabs}
\usepackage{multicol}
\usepackage{multirow}
\usepackage{tabularx}
\usepackage{enumitem}
\usepackage{hyperref}
\usepackage{placeins}
\usepackage{makecell}
\usepackage{stfloats}
\usepackage{float}

\usepackage[T1]{fontenc}
\usepackage[utf8]{inputenc}
\usepackage{csquotes}

\usepackage{microtype}

\usepackage{inconsolata}

\usepackage{graphicx}
\usepackage{subcaption}
\usepackage{arydshln}

\usepackage[most]{tcolorbox}
\usepackage{listings}
\usepackage{lstautogobble}
\usepackage{adjustbox}
\definecolor{DarkGreen}{RGB}{1,70,32}
\definecolor{paragraphbackground}{HTML}{A6CAEC}
\definecolor{promptbackground}{HTML}{DCEAF7}

\tcbset{
    backgroundbox/.style={
        colback=promptbackground,
        boxrule=1mm,
        arc=3mm,
        boxsep=.1pt,
        left=.5pt,
        right=.5pt,
        top=.5pt,
        bottom=.5pt,
        colframe=promptbackground
    },
    paragraphbox/.style={
        colback=paragraphbackground,
        boxrule=0.5mm,
        arc=1.5mm,
        auto outer arc,
        boxsep=.1pt,
        left=1pt,
        right=1pt,
        top=1pt,
        bottom=1pt,
        colframe=paragraphbackground
    }
}

\title{German4All – A Dataset and Model for Readability-Controlled Paraphrasing in German}

\author{Miriam Anschütz, Thanh Mai Pham, \\
\textbf{Eslam Nasrallah}, \textbf{Maximilian Müller}, \textbf{Cristian-George Craciun} \and \textbf{Georg Groh} \\
        Technical University of Munich \\ 
        \href{mailto:miriam.anschuetz@tum.de}{miriam.anschuetz@tum.de}, grohg@cit.tum.de
}

\begin{document}
\maketitle
\begin{abstract}
%The ability to paraphrase texts into different complexity levels enables generating flexible texts that can be tailored toward diverse reader groups. Therefore, we present German4All, the first German dataset of aligned paragraphs of multiple complexity levels, with more than 25k samples. The dataset was synthesized using GPT4 and thoroughly evaluated using human and LLM-based judges. Based on this dataset, we train an open-source multi-level paraphrasing model that supports tasks like simplification, paraphrasing, and complexification and marks a new milestone in reader-tailored simplification. With this new publicly available dataset, we foster further research in this direction.
The ability to paraphrase texts across different complexity levels is essential for creating accessible texts that can be tailored toward diverse reader groups. Thus, we introduce \textbf{German4All}, the first large-scale German dataset of aligned readability-controlled, paragraph-level paraphrases. It spans five readability levels and comprises over 25,000 samples. The dataset is automatically synthesized using GPT-4 and rigorously evaluated through both human and LLM-based judgments.\\
Using German4All, we train an open-source, readability-controlled paraphrasing model that %supports simplification, complexification, and general paraphrasing tasks. Our model 
achieves state-of-the-art performance in German text simplification, enabling more nuanced and reader-specific adaptations. We open-source both the dataset and the model to encourage further research on multi-level paraphrasing.

\end{abstract}

\section{Introduction}
%Why paragraph level?
%\citep{chi-different-complexity}
% https://aclanthology.org/2023.gem-1.18.pdf inherent level understanding in LLMs
% https://arxiv.org/abs/2409.20246 (Analysing Zero-Shot Readability-Controlled Sentence Simplification)
%- why complexification can be helpful (diverse input complexities -> better generalization, improve language understanding step-by-step)
Text simplification is typically approached as a standardized process, where an input text is simplified to a single, pre-defined complexity level—often determined by the model's training data. However, the audience for simplified language is highly diverse, including people with different reading proficiencies and backgrounds \citep{stajner-social-good}.
To better address this diversity, some languages such as German and Spanish differentiate between multiple simplification levels—for example, "plain language" and "easy-to-read" (DE: Leichte Sprache, ES: lectura fácil) \citep{maass-easy-plain-plus, madina-lectura-facil}. Despite this, existing simplification systems for German generally treat simplification as a one-size-fits-all task targeting a single output level.
\begin{figure}[ht]
    \centering
    \includegraphics[width=\linewidth]{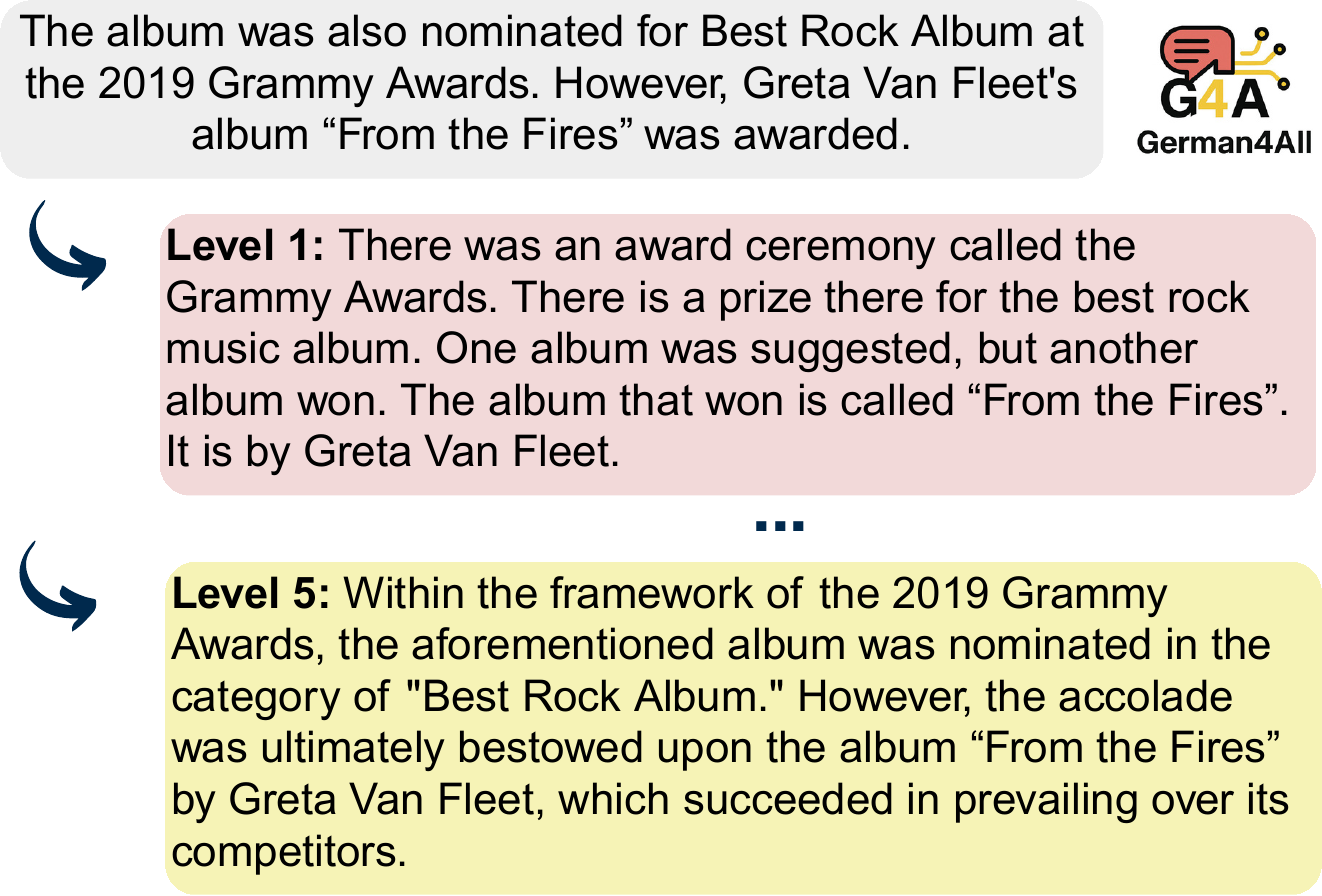}
    \caption{Example from our German4All dataset, translated to English. One input text is paraphrased into five versions at different complexity levels.}
    \label{fig:enter-label}
\end{figure}

%Text simplification is often performed as a standardized operation that simplifies an input text to a pre-defined complexity level, most often defined by the models' training data. However, the target group of simplified language is very diverse and includes people with different reading proficiencies and backgrounds \citep{stajner-social-good}. To account for this, languages like German or Spanish distinguish between different simplification levels, i.e., plain language and easy-to-read (DE: Leichte Sprache, ES: lectura fácil) \citep{maass-easy-plain-plus, madina-lectura-facil.
In contrast, English NLP research has explored methods for adapting simplification to multiple complexity levels \citep{chi-different-complexity, barayan-radability-control, farajidizaji-readability-control}. Yet, such fine-grained control remains underexplored in non-English contexts due to a lack of suitable resources. While powerful LLMs like GPT-4 can provide high-quality paraphrases in multiple languages, these models are harder to control for specific task settings. Moreover, users have to rely on API endpoints, introducing privacy concerns, or need to deploy large models in their environment \citep{Toshevska-finetuning-prompting}. Thus, we identify a need for smaller, task-specific models for readability-controlled paraphrasing.

To address this gap, we introduce German4All, the first large-scale, multi-level paraphrasing corpus for German. The dataset provides paraphrases of source texts across five distinct complexity levels. These levels are defined by their respective target group and range from people with reading difficulties to academic experts. For this paper, we refer to text complexity, readability, and simplification level as similar concepts.

The proposed dataset supports a range of tasks, including simplification, complexification, and readability-controlled paraphrasing. Moreover, the aligned paraphrases can be used in various settings, including iterative simplification or readability assessment. 
%Nevertheless, the existing German automatic text simplification systems treat the task as a one-size-fits-all solution to one pre-defined output level. While there exist approaches to paraphrase texts into various complexity levels in English \citep{chi-different-complexity, barayan-radability-control, farajidizaji-readability-control}, there is a lack of resources for non-English languages. Thus, we present the first large-scale multi-level paraphrasing corpus for German, containing paraphrases to five different complexity levels. It provides training data for simplification, paraphrasing, and complexification tasks, enabling fine-granular and customizable simplification with a better generalization to input with different complexity levels. Moreover, the aligned paraphrases can be used in various settings, including step-wise simplification, multi-level paraphrasing, or readability assessment.\\
Overall, our contribution can be summarized as follows:
\vspace{-0.3em}
\begin{itemize}[leftmargin=*]
\itemsep0em 
    \item We release \href{https://huggingface.co/datasets/tum-nlp/German4All-Corpus}{German4All}, a large-scale German dataset containing Wikipedia inputs with paraphrases to five different readability levels.
    % \item flexible usage (multi-level paraphrase, any-to-simple dataset, step-by-step simplification)
    \item While the corpus itself is synthesized using GPT4, we conduct a comprehensive evaluation—including an LLM-as-a-judge and feedback from 16 human participants—to validate its quality and usefulness for downstream tasks\footnote{\href{https://github.com/MiriUll/German4All}{https://github.com/MiriUll/German4All}}.
    \item We train a readability-controlled \href{https://huggingface.co/collections/tum-nlp/german4all-67477a7e9583754ed0bc3050}{German simplification model} on this dataset, which shows state-of-the-art performance compared to existing systems and reflects the styles of different versions of simplified language.
    % \item Both the dataset and model are released under MIT license.
\end{itemize}

\section{Related work}
% https://www.mdpi.com/2504-4990/7/3/68
\begin{figure*}[ht]
    \centering
    \includegraphics[width=\linewidth]{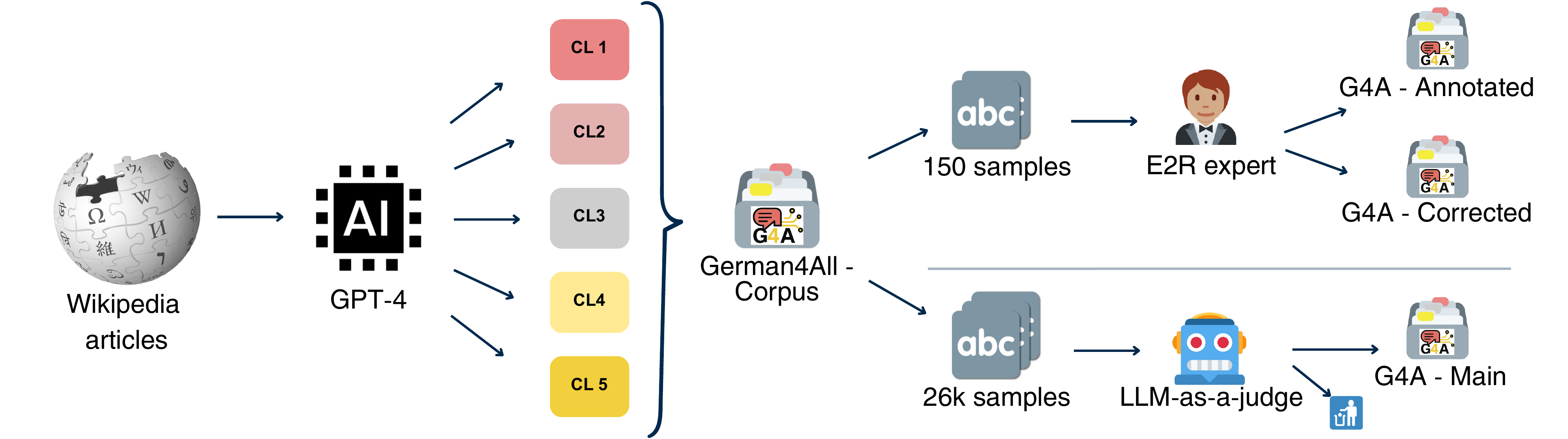}
    \caption{Overview of our dataset creation approach: We take paragraphs from Wikipedia and use GPT-4 to paraphrase them into five different complexity levels (CLs). Then, we manually curate a test set wth 150 samples, while automatically evaluating the main dataset with an LLM judge.}
    \label{fig:overview}
\end{figure*}
\paragraph{German text simplification} German exhibits different styles of simplified language. While simple/plain language (DE: "einfache Sprache") is a generally simpler version of standard German targeted to a broad audience, such as language learners \citep{din-es}, easy language (DE: "Leichte Sprache") is specifically tailored towards people with mental disabilities and reading deficiencies. As such, the simplifications in easy language are way stronger with shorter sentences and more guidance for the reader, e.g., through line breaks after every sentence \citep{din-spec-ls}. Many NLP resources exist for easy language \citep{schomacker-data, anschutz-language-models, toborek-simple-corpus}, simple language \citep{stodden-mt5-deplain, fruth-unsupervised-RF}, and other, more specific target groups like children \citep{aumiller-klexikon}. In addition, some works provide resources in multiple difficulty levels \citep{stodden-deplain}, most often following the Common European Framework of Reference for Languages (CEFR) \citep{council-cefr}. As such, \citet{spring-multi-level-simp} were the first to explore readability-controlled simplification for German. Despite these resources, there is a lack of large-scale and publicly available data for multi-level paraphrasing. We close this gap by providing a dataset with aligned paragraphs of different complexity levels. It not only supports simplification but also complexification to higher complexity levels. Complexification can be especially interesting for language learners who want to gradually improve their language proficiency. Moreover, input texts of various complexity can lead to a better generalization in simplification \citep{chi-different-complexity}.

% \subsection{Synthetic data / approach}
% veeeery similar but for Turkish: https://arxiv.org/pdf/2503.10675
% http://libstore.ugent.be/fulltxt/RUG01/003/209/858/RUG01-003209858_2024_0001_AC.pdf -> data = textbook and teacher-selected, complexity ~ school grade, slect texts from different grade books, NO alignment of data -> test llama-based models for their n-shot ability to create simplifications -> 95% of outputs are incomplete, LLms not ready for this task
% https://arxiv.org/pdf/2504.09394 SimpSyntEval: synthetic data for evaluation
% \cite{malik-gpt4-synthetic}: synthetic paragraphs with GPT-4, compare open-source models
% \cite{duran-synthetic-turkish-summary}: readability-controlled Turkish summaries, augment existing data with synthetic paraphrases (GPT) -> multiple readability levels of summaries, complexity ~ readability score

% \cite{spring-multi-level-simp}: sentence-level MT -> guide with target level tags, (MT pre-training task), sockeye architecture, use data from \cite{sauberli-benchmarking}, complexity ~ CEFR

% \cite{malik-gpt4-synthetic}: survey about can LLMs be complexity guided? EN only! -> GPT-4 successful, open-source bad -> RL-based training to improve open-source, complexity ~ CEFR

% \cite{kloser-synthetic}: see above

% \cite{almasi-alignment-drift}: non-English (Spanish)!, tutor-student chatbot, tutor with complexity-level guidance, differences between systems diminish over time -> not relevant for our setting, complexity ~ CEFR
\paragraph{Synthetic data generation}
We rely on synthetic data created with GPT-4 to compile a large-scale, multi-level dataset. Existing resources often rely on or extend other resources (e.g., the Simple German Corpus by \citealp{toborek-simple-corpus} is contained in the resources by \citealp{anschutz-language-models}). Thus, synthesizing a dataset enables the use of novel data and thus extends the diversity of available datasets. A similar approach was chosen by \citet{kloser-synthetic}, who synthesized complex data from existing human-created simplifications in German to obtain an aligned training corpus. In contrast to this, all our complexity levels are generated by GPT-4, giving us more control over the lexical diversity and the information preservation across all complexity levels. \citet{malik-gpt4-synthetic} benchmarked different open- and closed-source models for their ability to adhere to complexity guidance in English and found that only GPT-4 succeeds at this task without further fine-tuning. Similar results were reported by \citet{almasi-alignment-drift} who performed similar experiments for Spanish. While they find that GPT-4 loses guidance for longer chat histories, the initial answers comply with the complexity guidance in the system prompt.

% \paragraph{Readability-controlled simplification}

% \subsection{Controllable simplification}
% https://docnum.univ-lorraine.fr/public/DDOC_T_2023_0186_CRIPWELL.pdf#page=52.19
% https://dl.acm.org/doi/abs/10.1016/j.neucom.2024.127675                                                                             
% https://www.cambridge.org/core/journals/natural-language-engineering/article/how-do-control-tokens-affect-natural-language-generation-tasks-like-text-simplification/B252AD5B051CF5C764F5B3C9A2D46984
% https://sfb1102.uni-saarland.de/sfbunisb/uploads/2025/03/Thillainathan2025Controllable.pdf
% https://arxiv.org/abs/2309.12551

\section{Methodology}             
An overview of our data creation process is presented in \autoref{fig:overview}.
The inputs in our dataset are paragraphs from Wikipedia. For this, we selected the Wikipedia dump from December 2022 in the \href{https://huggingface.co/datasets/Cohere/wikipedia-22-12}{Cohere/wikipedia-22-12} dataset. This dataset contains Wikipedia paragraphs with at least 100 characters in multiple languages and sorts them by the number of views of their corresponding page. 
From this data, we randomly selected 26,665 samples from the 3 million most popular German paragraphs as our main data and 150 samples from the 1 million most popular German paragraphs as a test set, while assuring that the two subsets have no overlap. In the full dataset, the upper quartile of words per paragraph is 76 words. Hence, we excluded paragraphs with more than 80 words to ensure a diverse yet consistent paragraph length.
% Next, we prompted GPT-4 to create five different paraphrases across different complexity levels per input paragraph.
\subsection{Complexity levels}
% LS reduces content as well! LS+
Previous works by \citet{chi-different-complexity} or \citet{spring-multi-level-simp} have defined their complexity levels with existing CEFR levels \citep{council-cefr}. As such, the language levels of A1, A2, and partly B1 can be considered simple. Yet, these levels were mostly defined for language learners rather than accounting for language barriers of native speakers \citep{heine-gaf-ls}. Hence, the definitions of the different levels focus on the vocabulary and use cases at that level, e.g., introducing yourself in level A1. In contrast, people with learning disabilities have different backgrounds and require info about their everyday lives in an accessible language that goes beyond basic communication as a tourist.

Thus, we do not use CEFR levels but create our own complexity levels that are defined by the respective target groups and aligned with the literacy proficiency levels defined by the \citet[p. 64]{desjardins-oecd}. We distinguish five complexity levels between easy language for people with reading difficulties (level 1), commonly used language for the general public (level 3), and academic language for experts (level 5). The fine-grained definitions are presented in \autoref{sec:CL_levels}. Most input texts from Wikipedia are between levels 3 and 4, but as we cannot control for their level, they do not count towards the five levels provided.

\subsection{Synthetic data generation}
Empirical studies by \citet{manning-ki-tools} and \citet{barayan-radability-control} showed that ChatGPT with GPT-4 as backend can provide competitive and high-quality simplifications. Therefore, we used \textit{gpt-4-turbo-2024-04-09} via the OpenAI \href{https://platform.openai.com/docs/guides/batch}{Batch API} to create our complexity-aware paraphrases. % in 459 distinct batches. 
The model's system prompt was "\textit{You are an expert in adapting texts to different complexity levels.}". The more detailed user prompt is shown in \autoref{app:synt_data_prompt}. As suggested in the \href{https://help.openai.com/en/articles/6654000-best-practices-for-prompt-engineering-with-the-openai-api}{OpenAI engineering guide}, we structured the prompt into subsections. In these sections, we define our complexity levels, provide a 1-shot example and further details about the task, guide the model to pay attention to the previously described features of each of the complexity levels, and finally define the JSON output format. Our 1-shot example was randomly selected and manually paraphrased to all five complexity levels. Even though few-shot examples have been shown to improve the output quality \citep{malik-gpt4-synthetic}, we tried to find a balance between performance and cost due to the number of input tokens. Therefore, we only provided one example in the prompt.
\begin{table*}[ht]
    \centering
    \begin{tabularx}{\textwidth}{lXcc}\toprule
        \textbf{Split} & \textbf{Description} & \textbf{\#Samples} & \textbf{\#Paraphrases} \\
        \midrule
        \textit{main} & Main dataset, primarily for training & 25,459  & 5 (CL 1-5) \\
        \textit{corrected} & Manually corrected samples, extended with Leichte Sprache (LS) paraphrases & 150 & 6 (CL 1-5+LS)\\
        % &$\rightarrow$ training split & 100 & 6 \\
        % &$\rightarrow$ validation split & 20 & 6 \\
        % &$\rightarrow$ test split & 30 & 6 \\
        \textit{annotated} & Model outputs together with the manually corrected samples, annotated by the model's shortcomings & 132 & 2 \\
        \bottomrule
    \end{tabularx}
    \caption{Dataset splits and statistics. Each sample contains the original text together with different numbers of paraphrases at different complexity levels (CL).}
    \label{tab:data_statistics}
\end{table*}
\subsection{Data filtering}
Once the paraphrases were synthesized, we employed various automatic filtering steps to ensure a high overall quality of our dataset:
\begin{itemize}[leftmargin=*]
\itemsep0em 
    \item \textbf{Valid JSON format}: The output should be provided in a valid JSON format, and the paraphrases should be ordered by their complexity.
    \item \textbf{Out-of-vocabulary tokens}: We used spaCy \citep{honnibal-spacy} to flag samples with at least three consecutive tokens not in spaCy's vocabulary. These samples were manually reviewed and filtered for false positives.
    \item \textbf{Valid German text}: Using \href{https://github.com/Mimino666/langdetect}{langdetect}, we identified all samples containing other languages. Again, these samples were manually reviewed.
\end{itemize}
All samples that were still flagged as erroneous after the manual review were removed.
%Any sample that was correctly flagged by any of the filters was removed. 
In addition, we manually inspected random samples and discarded a minor proportion of them.
Ultimately, 26,337 out of 26,665 samples remained as the initial version of our German4All-Main dataset.

\subsection{Manual test data correction}\label{sec:annotated_corrected}
As outlined before, we sampled a subset of 150 distinct paragraphs from the Wikipedia corpus as our test set. This test set was manually revised and corrected by two native speakers, one of them being a Leichte Sprache expert. The 150 manually corrected samples form our German4All-Corrected subset. To enhance traceability and to facilitate further experiments, we annotated the changes that we performed during correction and stored these operations in the German4All-Annotated dataset. This dataset contains triplets of original texts, the original GPT-4 paraphrases to one specific complexity, and our corrected paraphrases. Our correction operations are categorized into six distinct operation types. %, and one category is true for a sample if we performed the respective operation. 
We distinguish between these operation annotations:
\begin{itemize}[leftmargin=*]
\itemsep0em
    \item \textit{removed\_info}: Whether we removed (potentially erroneous) information in the correction
    \item \textit{added\_info}: Whether we added information in the correction that was missing from the input
    \item \textit{corrected\_info}: Whether we fixed information in the correction 
    \item \textit{adjusted\_complexity}: Whether we corrected the language level to match the target complexity
    \item \textit{corrected\_language}: Whether we corrected language errors
    \item \textit{hallucination\_in\_origin}: Whether the original model output contains a hallucination    
\end{itemize}
%\textcolor{orange}{Observations from manual review?}

Furthermore, we manually created Leichte Sprache paraphrases for all samples in the corrected corpus. For this, we generated Leichte Sprache candidates with \href{https://easy-jon.de/}{EasyJon}, a free-to-use Leichte Sprache translation tool with an \textit{anthropic/claude-3.5-sonnet} backend \citep{Barbu-easyJon}. Then, the samples were manually rewritten by a German Easy Language expert.

Overall, this process returned three German4All subsets. An overview is provided in \autoref{tab:data_statistics}. While the main corpus is mainly useful for training, the corrected corpus can serve as a gold-standard evaluation dataset or be used for RLHF approaches.

\section{Dataset evaluation}
Our synthesized data contains more than 25k samples with five complexity level paraphrases each, resulting in a dataset of more than 125k text pairs. Due to this size, a human review of all samples would be infeasible. Therefore, we investigated two different evaluation angles. First, we randomly selected samples from the corpus and performed a human evaluation on these samples. Then, we extended our evaluation to the full corpus by using an LLM-as-a-judge.
\subsection{Human evaluation}
%n participants, dataset versions to increase coverage while still considering subjective differences
%low agreement in some questionnaire versions, probably due to small data size
For the human evaluation, we selected 15 samples with five paraphrases each (=75 text pairs in total) from the Main subset at random. These samples were split into five groups with three samples and 15 original-paraphrased text pairs (3 samples * 5 complexity levels) per group. The paraphrase pairs were presented one by one, grouped by their original text and sorted by their complexity level. For each text pair, we asked the following questions with the answer options in parentheses. The evaluation was originally conducted in German, but we translated it here:
\begin{itemize}
\itemsep0em
    \item[Q1] The paraphrase reflects the content of the original text ... [\textit{incorrectly} | \textit{approximately} | \textit{correctly}]
    \item[Q2] How often were pieces of information from the original text omitted in the paraphrase? [\textit{never} | \textit{seldom} | \textit{sometimes} | \textit{often}]
    \item[Q3] How often were additional pieces of information added in the paraphrase that were not present in the original text? [\textit{never} | \textit{seldom} | \textit{sometimes} | \textit{often}]
    \item[Q4] Skip this question if you selected ”never” in the previous question. What types of additional information were added? Multiple options can be selected.\newline [\textit{embellishment} | \textit{explanations or definitions} | \textit{factually incorrect information} | \textit{factually correct information} | \textit{other}]
    \item[Q5] The paraphrase for the desired difficulty level is ... [\textit{too easy} | \textit{a bit too easy} | \textit{appropriate} | \textit{a bit too complicated} | \textit{too complicated}]
\end{itemize}

In total, 16 native German speakers participated in our human evaluation. They participated voluntarily and received no financial compensation. Each participant was assigned to one of the five sample groups, so that we had at least three participants per group. We divided the participants into groups to balance the workload of a single participant while receiving multiple feedback for as many samples as possible, thus increasing the coverage of our evaluation. With this setup, participants needed $\pm$20 minutes to complete their evaluation, and all indeed completed the survey. 
\begin{table}[ht]
    \centering
    \small
    \begin{tabular}{lccccc|c}\toprule
        & \textbf{G1} & \textbf{G2} & \textbf{G3} & \textbf{G4} & \textbf{G5} & \textbf{Mean}\\ \midrule
         % \textbf{Q1} & 0.31 \footnotesize{$\pm$ 0.14} & \\
         % \textbf{Q2} & 0.47 \footnotesize{$\pm$ 0.32} & \\
         % \textbf{Q3} & 0.40 \footnotesize{$\pm$ 0.27} & \\
         % \textbf{Q5} & 0.09 \footnotesize{$\pm$ 0.17} & \\
         \textbf{Q1} & 0.53 & 0.29 & 0.26 & 0.17 & 0.28 & 0.31\\
         \textbf{Q2} & 0.73 & 0.56 & -0.05 & 0.72 & 0.38 & 0.47\\
         \textbf{Q3} & 0.54 & 0.18 & 0.67 & 0.05 & 0.56 & 0.40\\
         \textbf{Q5} & -0.16 & 0.04 & 0.13 & 0.16 & 0.29 & 0.09\\
         \textbf{Q5 Tol.} & -0.25 & 0.2 & 1.0 & 1.0 & 0.31 & 0.45 \\
        \bottomrule
    \end{tabular}
    \caption{Inter-annotator agreement measured by Krippendorff's $\alpha$ for questions 1-3 and 5 across annotator groups (G1-G5).  For Q5 (complexity level), we also report the agreement with a $\pm 1$ tolerance.}
    \label{tab:human_agree}
\end{table}
\begin{figure*}
    \centering
    \begin{subfigure}[c]{\linewidth}
        \includegraphics[width=\linewidth]{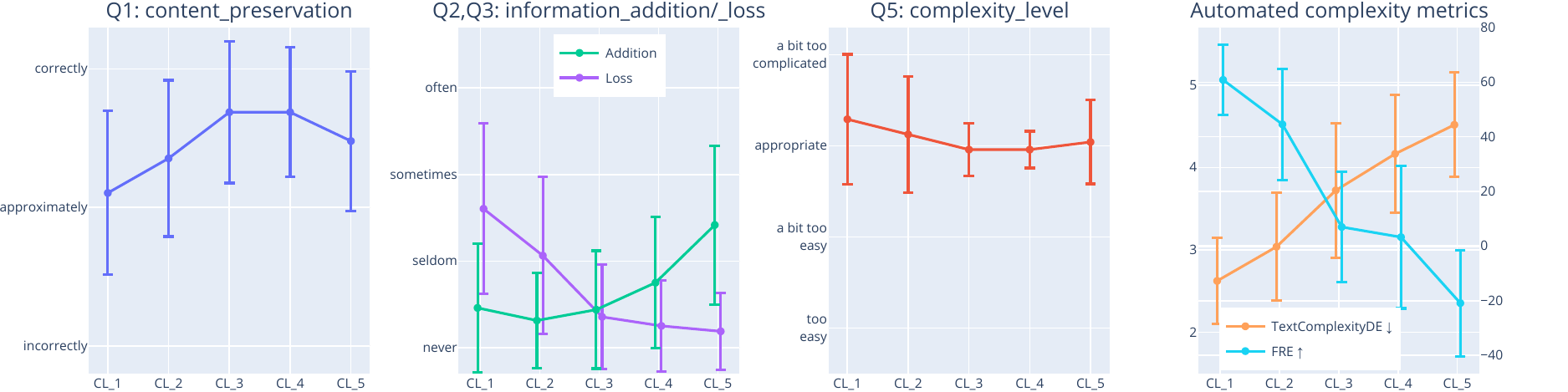}
        \subcaption{Human evaluation results and automated metrics for the respective subset (75 original-paraphrased pairs).}
        \label{fig:human_eval}
    \end{subfigure}
    \begin{subfigure}{\linewidth}
        \includegraphics[width=\linewidth]{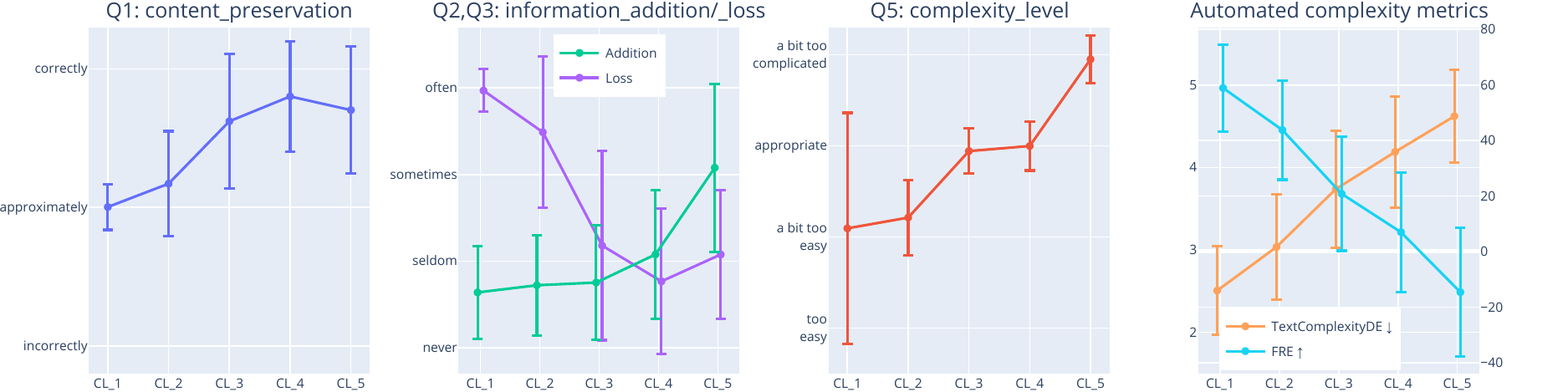}
        \subcaption{LLM judge evaluation and automated metrics on the full main split (131.365 original-paraphrased pairs).}
        \label{fig:LLM_eval}
    \end{subfigure}
    \caption{Evaluation results across questions 1-3 and 5, grouped by the texts' complexity level (CL). The rightmost plot shows two automatic readability measures, TextComplexityDE (\citealp[lower means easier]{anschutz-TextComplexityDE}) and Flesh readability scores (\citealp[higher means easier]{Amstad-FRE}), to automatically measure the text complexity.}% The statistics are calculated among all answers across all sample groups.}
    \label{fig:questionaire_eval}
\end{figure*}

\autoref{tab:human_agree} shows the agreement scores for the different evaluation groups using \href{https://github.com/grrrr/krippendorff-alpha}{Krippendorff's $\alpha$} \citep{krippendorff-alpha}. Considering the low number of annotators, the content-related questions show a decent average agreement. Yet, the complexity level question has a mixed result. Since text complexity is a very subjective measure, we also calculated the agreement with a tolerance of  one level (e.g., "\textit{too easy}" and "\textit{a bit too easy}" are considered to be an agreement). Still, the agreement for the first evaluation group seems to be very low. However, all three annotators give exactly the same score for a sample in 66\% of all pairs, so this is mostly an artifact of the agreement score calculation. Overall, we consider the agreement to be good enough to deduce trends and conclusions from the evaluation.

\autoref{fig:human_eval} shows the answers for Q1-3 and Q5. The content is best preserved for complexity levels 3 and 4, while levels 1 and 2 show the highest rates of information loss. This is expected for simplifications that leave out (minor) details to improve understanding \citep{trienes-infolossQA}, and thus, a desired feature of the dataset. In contrast, the higher complexity levels add the most information. When looking at the answers from Q4, the most information added in these complexity levels is embellishments and facts. For CL\_1, this shifts toward explanations or definitions, which improve the readability of a text. Looking at the complexity levels, we compare the human answers with the Flesh readability ease (FRE) score by \citet{Amstad-FRE}, where a higher FRE score indicates a better readability. In addition, we used the open-source text complexity prediction model for German texts by \citet{anschutz-TextComplexityDE}, trained on the TextComplexityDE dataset \citep{Naderi-dataset}. Here, the scores range from 1 (easiest) to 7. 
Most of the texts seem to be appropriate for their respective complexity level, with the strongest human disagreement for the edge levels 1 and 5. Text complexity is a subjective measure, and thus, it can be expected that these levels show a higher answer variation.

Overall, the human evaluation on the randomly selected subset shows that our dataset is of high quality and that the generated paraphrases exhibit the features of their respective complexity levels.
\subsection{LLM-as-a-judge}\label{sec:llm_judge}
% https://arxiv.org/pdf/2406.12624v1#page=0.84
The human evaluation only covers a small proportion of our overall dataset. To acquire a large-scale evaluation of all data samples, we employ an LLM-as-a-judge as automatically evaluating simplification with LLMs has shown good correlation with human judgements in previous work \citep{liu2025-llm-judge-simplification}.
We selected \textit{\href{https://huggingface.co/google/gemma-3-27b-it}{gemma-3-27B-it}} as the judge model due to its model size, good multilingual capabilities, and good performance in preliminary experiments (see \autoref{app:llm_ablation_models}). Our setup reuses the questions and evaluation criteria from the human evaluation. However, every pair of original and paraphrased texts is provided individually without the context of the other paraphrase levels, resulting in $5*26,337=131,685$ evaluation pairs. The prompt is written in German to avoid code-switching and provides a one-shot example. We utilize structured outputs to force the model to output a parseable JSON output. The full prompt with the task description and further guidance for the model is presented in \autoref{app:llm_judge_prompt}.

To evaluate and improve our LLM judge, we selected the Annotated subset and tested whether the judge is sensitive to our criteria and correctly annotates the faulty and correct samples. For this, we provided the original text and the original model outputs to the LLM judge, and we transferred the textual answers into a numerical scale and 
performed Spearman correlation tests between our edit operation annotations (compare \autoref{sec:annotated_corrected}) and the judge outputs. For the complexity level, we considered the answers \textit{a bit too easy} and \textit{a bit too complex} as suitable and not requiring an adjustment of the text complexity. 
We found significant correlations between multiple aspects, including the information\_addition (LLM judge) and human hallucination annotations, or complexity\_level (LLM judge)  and the human complexity adjustments. The full correlation analysis, as well as a comparison between different judge models, is presented in \autoref{app:llm_ablation_models}. In addition, we provide an example from the dataset and its LLM judge annotations in \autoref{app:judge_example}.
%The annotation pairs with significant correlations (p-value $\leq 0.01$) are presented in \autoref{tab:judge_annotated_correlations}.
We concluded that the LLM judge can successfully detect errors in the synthesized outputs, and thus, used it to filter the Main dataset.
% \begin{table}[ht]
%     \centering
%     \begin{tabularx}{\linewidth}{p{2.3cm}Xl}\toprule
%         \textbf{\makecell[l]{Human\\ operation}} & \textbf{\makecell[l]{LLM Judge\\criterion}} & \textbf{Stat} \\
%         \midrule
%         % content\_preservation & removed\_info & 0.31 \\
%          corrected\_info & content\_preservation & -0.27\\
%         removed\_info & information\_addition & 0.51 \\
%         hallucination \_in\_origin & information\_addition & 0.35 \\
%         adjusted \mbox{\_complexity} & complexity\_level & 0.34 \\
%         \bottomrule
%     \end{tabularx}
%     \caption{Selected edit operations in the German4All-Annotated dataset with a significant correlation (p-value $\leq 0.01$) with the respective LLM judge criteria. The stat value shows the strength of the correlation.}
%     \label{tab:judge_annotated_correlations}
% \end{table}

\autoref{fig:LLM_eval} shows the LLM judge's evaluation on the full dataset. In general, the curves look similar to the results of our human evaluation. However, the LLM judge seems to be stricter with the information loss in the lower complexity levels, resulting in only an \textit{approximate} content preservation. As discussed before, text simplification focuses only on the most important information to improve understanding. The strongest differences are evident for the perceived text complexity, as the LLM seems to struggle with the definition of an appropriate complexity level and rather rates the absolute complexity of the text. Therefore, the samples in the lowest complexity level are annotated as \textit{(a bit) too easy} and the samples in the highest complexity level as \textit{a bit too complicated}. Since these are our most extreme levels, we expect them to be as easy/complicated as possible. In addition, the automated metrics show that our samples indeed have the target complexity.

Based on the LLM judge's feedback, we further filtered the main split and automatically removed (a number of) samples where:
\vspace{-3pt}
\begin{itemize}[leftmargin=*]
\itemsep0em
    \item the content\_preservation is \textit{incorrect} (340).
    \item the complexity\_level is \textit{too easy} for all complexity levels except CL\_1 (6).
    \item the complexity\_level is \textit{too complex} for all complexity levels except CL\_5 (0). % filter_ids_zu_kompliziert = 0 samples
    \item the type of added information is \textit{factually incorrect information} (83) or \textit{other} (5), and \textit{factually correct information} for all except CL\_5 (67).
    \item the information\_loss is \textit{often}, but only for samples in CL\_5 (403). \vspace{-3pt}
\end{itemize}
Some samples were flagged by more than one criterion, resulting in 814 removed samples in total, and our final German4All-Main dataset with 25,459 samples.
The LLM judge annotations are uploaded in our GitHub repository, so users can create their own filters if needed.

\section{Model distillation}
\begin{figure*}
    \centering
    \includegraphics[width=\linewidth]{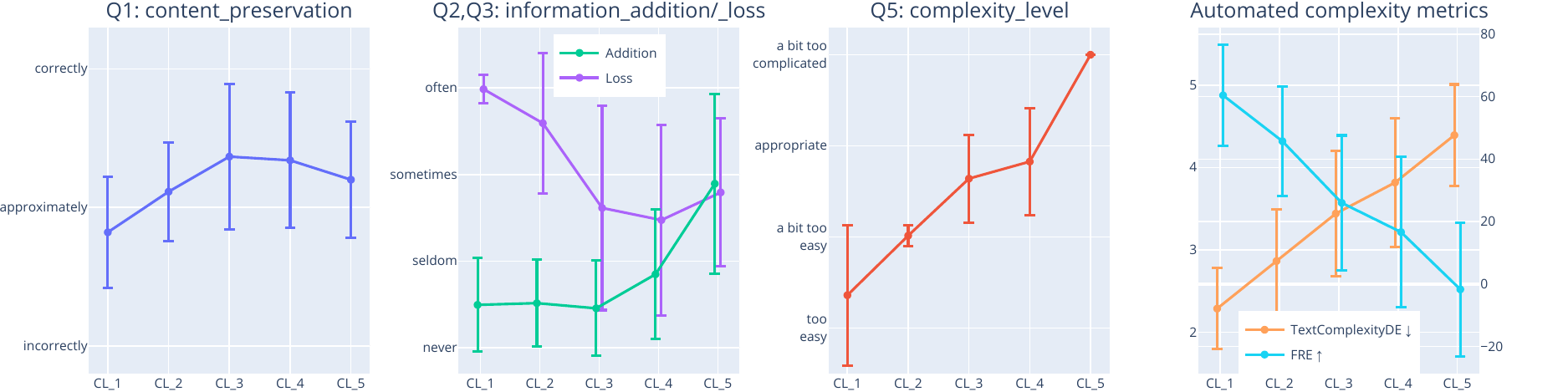}
    \caption{LLM judge evaluation of the distilled model's predictions on the German4All-Corrected test set.}
    \label{fig:single_model_eval}
\end{figure*}
% custom-decoder, DEplain, Llama 3.3
% https://huggingface.co/frhew/sigdial_ft_a2 -> https://huggingface.co/hiig-ai-lab/simba_best_092024 v
% https://github.com/MSLars/German-Text-Simplification v
% https://github.com/LFruth/unsupervised-german-ts/tree/master <- missing
%Simba: fine-tuned Llama-3-8B-Instruct
The ultimate motivation for our dataset is to create a smaller model that can perform the task of readability-controlled paraphrasing without having to rely on large or expensive models. Therefore, we have inserted and trained LoRA layers \citep{hulora} for a \href{https://huggingface.co/google/flan-t5-xl}{flan-t5-xl} model \citep{chung-flan-t5} that can be inferenced on consumer-sized graphic cards with 12GB of VRAM. We also experimented with different base models such as Llama3.1 8B \citep{dubey-llama3} and the German-specific Lämmlein 7B \citep{pfister-lammlein}. However, these models exhibited many grammatical errors even before any fine-tuning. Thus, we chose Flan-T5 as base, even though it is already fairly old.

We trained it using a random train:val-80:20 split of the German4All-Main corpus. For every level, we took not only the original input but also the other complexity levels' paraphrases as inputs. This increased our training data by five and leads to a better model generalization as the model sees different styles and complexities as inputs. 

For our prompt design, we use the same German complexity level descriptions as for the dataset creation (\autoref{sec:CL_levels}). These descriptions are followed by the task "\textit{Paraphrase the following text to level \{level\}. \{input\_text\}}" (translated here).
\subsection{Evaluating model performance}

To evaluate the performance of our distilled model, we benchmarked its performance on the German4All-Corrected test set. Therefore, we created predictions for all complexity levels of all 150 samples, resulting in 750 predictions. Then, we employed the LLM judge and automatic readability metrics from \autoref{sec:llm_judge} to evaluate them. The results are presented in \autoref{fig:single_model_eval}: The model successfully learns the characteristics of the different complexity levels and outputs appropriate texts. Yet, from the content perspective, we observe a stronger deviation and a higher content loss compared to the original dataset. To further investigate this, we manually reviewed 75 model predictions (15 per CL). We find that the syntax and style of the samples are of very high quality and match the expected style of the respective level. However, we also found issues with fluency and minor grammatical errors. Moreover, the samples in the higher complexity levels often contain hallucinations, aligning with observations from the LLM judge's \textit{information\_addition} criterion. Overall, we find the paraphrases succeed at presenting an input text at different language levels.

\subsection{Benchmarking simplification performance}
\begin{table*}[ht]
    \centering
    \small
    \begin{tabular}{lccccccc}\toprule
         \textbf{Model} & \textbf{BLEU}$\uparrow$ & \textbf{SARI}$\uparrow$ &  \textbf{BS\_F1}$\uparrow$ & \textbf{FRE}$\uparrow$ & \textbf{Compression}$\downarrow$ & \textbf{Sent. splits}$\uparrow$ & \textbf{Copies}$\downarrow$ \\ \midrule
         \multicolumn{8}{c}{\textit{DEplain-web-sent \citep{stodden-deplain}}} \\References & - & - & - & 75.56 & 0.94 & 1.87 & 0.0 \\\hdashline \\[-1.7ex]
        mBART-DEplain-APA+web & \textbf{17.99} & 34.07 & \textbf{0.43} & 68.41 & 0.85 & 1.16 & 0.14\\
        mt5-SGC & 3.0 & 37.02 & 0.29 & 74.45 & \textbf{0.5} & 0.95 & 0.01\\
        German4All-level1 (ours) & 5.58 & 37.85 & 0.32 & \textbf{84.81} & 0.96 & \textbf{2.3} & \textbf{0.00}\\
        German4All-level2 (ours) & 7.67 & \textbf{38.05} & 0.36 & 77.27 & 1.01 & 1.79 & \textbf{0.00}\\ \midrule
         \multicolumn{8}{c}{\textit{Simple German Corpus \citep{toborek-simple-corpus}}} \\References & - & - & - & 64.54 & 1.26 & 2.15 & 0.0\\\hdashline \\[-1.7ex]
        mBART-DEplain-APA+web & \textbf{6.69} & 28.68 & \textbf{0.32} & 44.23 & 1.62 & 1.29 & 0.17\\
        mt5-SGC & 4.67 & \textbf{43.68} & \textbf{0.32} & 57.85 & \textbf{0.66} & 1.02 & 0.03\\
        German4All-level1 (ours) & 3.2 & 41.09 & 0.27 & \textbf{79.2} & 1.18 & \textbf{2.11} & \textbf{0.00}\\
        German4All-level2 (ours) & 4.36 & 39.41 & 0.29 & 65.53 & 1.37 & 1.62 & 0.01\\ \midrule
         \multicolumn{8}{c}{\textit{GEOlino \citep{Mallinson-geolino}}} \\References & - & - & - & 68.18 & 0.95 & 1.32 & 0.36 \\\hdashline \\[-1.7ex]
        mBART-DEplain-APA+web & \textbf{55.35} & \textbf{44.28} & \textbf{0.77} & 67.23 & 0.97 & 1.09 & 0.28\\
        mt5-SGC & 12.6 & 29.17 & 0.49 & 75.26 & \textbf{0.75} & 0.96 & 0.04\\
        German4All-level1 (ours) & 12.46 & 29.18 & 0.44 & \textbf{83.54} & 1.13 & \textbf{1.89} & \textbf{0.00} \\
        German4All-level2 (ours) & 19.86 & 34.05 & 0.54 & 75.63 & 1.2 & 1.5 & 0.01\\
        \midrule
         \multicolumn{8}{c}{\textit{TextComplexityDE \citep{Naderi-dataset}}} \\References & - & - & - & 51.64 & 0.95 & 2.08 & 0.0 \\\hdashline \\[-1.7ex]
        mBART-DEplain-APA+web & \textbf{17.75} & 37.37 & \textbf{0.5} & 45.43 & 0.74 & 1.31 & 0.06\\
        mt5-SGC & 1.52 & 33.51 & 0.27 & 65.12 & \textbf{0.34} & 0.96 & \textbf{0.00}\\
        German4All-level1 (ours) & 4.28 & 35.43 & 0.29 & \textbf{81.39} & 0.77 & \textbf{3.04} & \textbf{0.00}\\
        German4All-level2 (ours) & 9.33 & \textbf{40.56} & 0.42 & 67.11 & 0.81 & 2.18 & \textbf{0.00}\\
        \bottomrule
    \end{tabular}
    \caption{Performance comparison on four German simplification datasets. We follow the approach by \citet{stodden-mt5-deplain}. Thus, we evaluated the outputs in terms of the reference-based metrics BLEU \citep{papineni-bleu}, SARI \citep{Xu-SARI}, and BERTscore \citep{Zhang-BERTScore}, as well as with the linguistic annotations FRE, compression rate, number of sentence splits, and number of exact copies. The best value per dataset is bolded.}
    \label{tab:simp_benchmarks}
\end{table*}
% DEplain, 20 minute <- no, simple-german corpus (toborek), Klaper?
% https://github.com/mlai-bonn/Simple-German-Corpus
% https://huggingface.co/datasets/MSLars/erlesen_synth_v1
In addition to judging the quality of our model's outputs, we benchmarked and compared its simplification performance compared to other German ATS systems on existing text simplification datasets. For this, we used the EASSE-DE evaluation suite \citep{stodden-easse-de}.
First, we compared the performance on our test set with the latest German ATS models. For this, we compare these four models: \href{https://huggingface.co/MSLars/erlesen-leo-7b}{erlesen-leo-7b} \citep{kloser-synthetic}, \href{https://huggingface.co/DEplain/trimmed_mbart_sents_apa_web}{mBART-DEplain-APA+web} \citep{stodden-deplain}, \href{https://huggingface.co/DEplain/mt5-simple-german-corpus}{mt5-simple-german-corpus (SGC)} \citep{stodden-mt5-deplain}, and \href{https://huggingface.co/hiig-ai-lab/simba_best_092024}{Simba} \citep{hewett-simba}.
The full evaluation results are presented in \autoref{app:g4a_benchmark}. Our model outperforms the other systems in all evaluation criteria except the compression strength. More interestingly, though, is that our model's target complexity outputs match the characteristics of the respective references, i.e., to-level-1 wins for the level 1 references, but to-level-2 wins for the level 2. This shows that our model indeed distinguishes between different complexity levels.

Finally, we evaluated our model on publicly available sentence- or paragraph-level datasets, following the approach by \citet{stodden-mt5-deplain}. \autoref{tab:simp_benchmarks} shows the performance when prompting our model to output texts in complexity levels 1 or 2. We compare them with two models that performed best in terms of SARI and FRE. The full EASSE-DE reports can be found on git\footnote{\url{https://github.com/MiriUll/German4All/tree/main/easse-de_reports}}. Our models achieve SOTA SARI  scores and the highest number of sentence splits, showing a consistent performance on all datasets. Yet, they fall short in terms of meaning overlap metrics like BLEU and BERTscore. EASSE-DE outputs the samples with the worst relative score compared to the other models. We manually reviewed these samples and found that often, the input and reference data are noisy, and our model outputs are quite good, matching the expected styles of levels 1 and 2. An example for this is the following sample from the DEplain-web-sent corpus \citep{stodden-deplain}:
\paragraph{Input:} "Ein beliebter Badeort mit vielen Stränden und Wassersport ist Deutschlands größter See , der Bodensee , ganz im Süden ." ({\small \textit{EN: A popular seaside resort with many beaches and water sports is Germany's largest lake, Lake Constance, in the far south.})}\\
\textbf{German4All-level1}: "In Deutschland gibt es einen großen See . Der Bodensee ist sehr groß . Viele Menschen gehen dort zum Baden ." ({\small \textit{EN: There is a large lake in Germany . Lake Constance is very big . Many people go there to swim .})}\\
\textbf{German4All-level2}: "Der Bodensee ist ein großer See in Deutschland . Er liegt im Süden . Dort gibt es viele Strände und viele Möglichkeiten , Wassersport zu machen ." ({\small \textit{EN: Lake Constance is a large lake in Germany . It is located in the south . There are many beaches and many opportunities to do water sports .})}

Both models provide a high-level simplification of the input and preserve most of the content. The level 1 simplification is easier than level 2, with stronger content compression and shorter sentences. Nevertheless, both simplifications receive low automatic scores due to the poor reference. This problem with automatic meaning preservation metrics was also reported by \citet[Section 5]{barayan-radability-control} and will probably remain unsolved until a multilingual paraphrase-aware metric is developed. 

% Original
% 2017 gehörte der Sänger und Songwriter Mark Forster mit „ Chöre “ zu den erfolgreichsten deutschsprachigen Musikern .
% DEplain_trimmed_mbart_sents_apa_web.txt
% Im Jahr 2017 gehörte der Sänger und Songwriter Mark Forster mit Chöre zu den erfolgreichsten deutschsprachigen Musikern .
% mt5_simple-german-corpus.txt
% Mark Forster ist der Sänger und Sänger von Deutschland .
% German4All_level_1.txt
% Mark Forster ist ein Sänger . Er hat ein Lied namens ' Chöre ' gemacht . Das Lied ist sehr beliebt .
\section{Conclusion}
In this paper, we present German4All, the first aligned multi-level paraphrasing dataset in German. With more than 25k samples and a Flan-T5 model trained on this data, our work enables large-scale research on text simplification, paraphrasing, and readability assessment—going beyond what was previously possible for German.
\FloatBarrier
\section*{Limitations}
GPT4 has shown impressive performance and high agreement with human annotators in readability ranking tasks \citep{engelmann-arts} or when generating simplifications \citep{kloser-synthetic}. While we tried to give an exhaustive evaluation of our dataset with 16 native speakers and an LLM-as-a-judge, the dataset was synthesized using an LLM. Thus, it could still contain errors or biases, and users of the dataset must be aware of these limitations. Moreover, the human annotators have a considerably high degree of education and can't be considered people from the easy or plain language target groups. This restriction was necessary so that the annotators could also evaluate the samples in the higher complexity levels 4 and 5. Nevertheless, it results in a lack of target group integration that is recommended for readability evaluations \citep{sauberli-eye-tracking} and has been done for other works in simplification \citep{gao-target-eval,anschutz-etal-2024-images}.

In terms of content, we focused on data from Wikipedia due to its permissive license. This results in samples that are very descriptive and have an explanatory nature. Yet, we not only took paragraphs from the abstracts, but from the full articles. Thus, our data also contains paragraphs that describe happenings and the life lines of people. These samples can be considered similar to news articles, and thus, our dataset can be used for paraphrases in different domains and use cases. The Wikipedia samples could have been part of GPT-4's training data and thus contaminated. However, we don't benchmark GPT-4, but use it to create our dataset. Thus, we don't see any problems with this exposure.

Finally, our simplifications show a very good structural understanding of simplified language and introduce sentence splits and rewritings where necessary. However, while very complex terms are often removed, the level of factual term explanations and guiding the user to interpret the content could be increased \citep{hewett-elaborative}. However, to ensure factual correctness in these definitions, we did not include this angle of elaborative simplification in our dataset.

% No CoT prompting

% complexity rating by non-target group no good measure (cite säuberli eye tracking), but automated metrics and statistics show success in complexity adaption
% also: must evaluate all samples of all CLs -> target group evaluation not suitable

% fully synthetic data, not all samples manually reviewed!!
% GPT-4 shows very high agreement with human annotators in a readability ranking task: 

% focus only on Wikipedia -> extend to other or more specific domains (but many samples very descriptive of what happened, so could transfer to news domain)

% enhance with elaborative simplification

\section*{Ethical considerations}
Readability-controlled paraphrasing is an effort to tailor texts to the needs of specific readers. Thus, it tries to overcome the limitations of text simplification that assumes a homogeneous target group and only provides standardized simplifications. As such, these single simplifications may still be too complex for some readers, while others might find the level of simplification discriminating \citep{maass-easy-plain-plus}. Therefore, our resources aim to help reduce discrimination while increasing true accessibility. 

However, synthesizing a dataset with LLMs like GPT-4 might introduce biases into our dataset that we could not control for. As such, the LLM outputs reflect the biases and limitations of GPT-4's training data. Moreover, our input data relies on Wikipedia, a user-generated information platform with potentially erroneous content. Thus, users of our dataset and model should be aware of potentially wrong or outdated information and always evaluate the outputs against other sources.

\section*{Data availability statement}
All experimental results that were discussed in this work are publicly available on our Git repository or on Huggingface. This includes the different versions of our dataset, results from the human, LLM judge, and EASSE-DE evaluation, and all model predictions.

\section*{Acknowledgments}
Our paper uses closed-source models from OpenAI. To create and evaluate our corpus, we spent approximately 500\$.\\
This paper is based on joint work in the context of Mai Pham's master’s thesis \cite{Pham-MA-German4All}. In addition, we thank Robert Schauer, Maximilian Eckert, and Ricardo Ebene for their preliminary model fine-tuning results.\\
This research has been funded by the German Federal Ministry of Research, Technology and Space (BMFTR) through grant 01IS23069 Software Campus 3.0 (Technical University of Munich) as part of the Software Campus project \enquote{LIANA}.\\
The authors used an AI writing assistant in the form of ChatGPT to improve formulations and phrasing in the paper. Yet, no novel text was generated, and all corrections were thoroughly revised by the authors.
\bibliography{custom}

\begin{thebibliography}{49}
\providecommand{\natexlab}[1]{#1}

\bibitem[{Almasi and Kristensen-McLachlan(2025)}]{almasi-alignment-drift}
Mina Almasi and Ross Kristensen-McLachlan. 2025.
\newblock \href {https://doi.org/10.18653/v1/2025.bea-1.6} {Alignment drift in
  {CEFR}-prompted {LLM}s for interactive {S}panish tutoring}.
\newblock In \emph{Proceedings of the 20th Workshop on Innovative Use of NLP
  for Building Educational Applications (BEA 2025)}, pages 70--88, Vienna,
  Austria. Association for Computational Linguistics.

\bibitem[{Amstad(1978)}]{Amstad-FRE}
Toni Amstad. 1978.
\newblock \emph{Wie verst{\"a}ndlich sind unsere Zeitungen?}
\newblock Ph.D. thesis, Universit{\"a}t Z{\"u}rich.

\bibitem[{Ansch{\"u}tz and Groh(2022)}]{anschutz-TextComplexityDE}
Miriam Ansch{\"u}tz and Georg Groh. 2022.
\newblock \href {https://aclanthology.org/2022.germeval-1.4/} {{TUM} social
  computing at {G}erm{E}val 2022: Towards the significance of text statistics
  and neural embeddings in text complexity prediction}.
\newblock In \emph{Proceedings of the GermEval 2022 Workshop on Text Complexity
  Assessment of German Text}, pages 21--26, Potsdam, Germany. Association for
  Computational Linguistics.

\bibitem[{Ansch{\"u}tz et~al.(2023)Ansch{\"u}tz, Oehms, Wimmer, Jezierski, and
  Groh}]{anschutz-language-models}
Miriam Ansch{\"u}tz, Joshua Oehms, Thomas Wimmer, Bart{\l}omiej Jezierski, and
  Georg Groh. 2023.
\newblock \href {https://doi.org/10.18653/v1/2023.findings-acl.74} {Language
  models for {G}erman text simplification: Overcoming parallel data scarcity
  through style-specific pre-training}.
\newblock In \emph{Findings of the Association for Computational Linguistics:
  ACL 2023}, pages 1147--1158, Toronto, Canada. Association for Computational
  Linguistics.

\bibitem[{Ansch{\"u}tz et~al.(2024)Ansch{\"u}tz, Sylaj, and
  Groh}]{anschutz-etal-2024-images}
Miriam Ansch{\"u}tz, Tringa Sylaj, and Georg Groh. 2024.
\newblock \href {https://doi.org/10.18653/v1/2024.tsar-1.4} {Images speak
  volumes: User-centric assessment of image generation for accessible
  communication}.
\newblock In \emph{Proceedings of the Third Workshop on Text Simplification,
  Accessibility and Readability (TSAR 2024)}, pages 27--40, Miami, Florida,
  USA. Association for Computational Linguistics.

\bibitem[{Asghari et~al.(2024)Asghari, Richter, Hewett, Wunderlich, and
  Züger}]{hewett-simba}
Hadi Asghari, Christopher Richter, Freya Hewett, Larissa Wunderlich, and
  Theresa Züger. 2024.
\newblock \href {https://publicinterest.ai/tool/simba?lang=en} {Simba:
  Ai-powered text simplification}.

\bibitem[{Aumiller and Gertz(2022)}]{aumiller-klexikon}
Dennis Aumiller and Michael Gertz. 2022.
\newblock \href {https://aclanthology.org/2022.lrec-1.288/} {Klexikon: A
  {G}erman dataset for joint summarization and simplification}.
\newblock In \emph{Proceedings of the Thirteenth Language Resources and
  Evaluation Conference}, pages 2693--2701, Marseille, France. European
  Language Resources Association.

\bibitem[{Barayan et~al.(2025)Barayan, Camacho-Collados, and
  Alva-Manchego}]{barayan-radability-control}
Abdullah Barayan, Jose Camacho-Collados, and Fernando Alva-Manchego. 2025.
\newblock \href {https://aclanthology.org/2025.coling-main.452/} {Analysing
  zero-shot readability-controlled sentence simplification}.
\newblock In \emph{Proceedings of the 31st International Conference on
  Computational Linguistics}, pages 6762--6781, Abu Dhabi, UAE. Association for
  Computational Linguistics.

\bibitem[{Barbu(2024)}]{Barbu-easyJon}
Paul-Gerhard Barbu. 2024.
\newblock Entwicklung einer anwendung zum Übersetzten von texten in
  leichter/einfacher sprache mithilfe von large language models (llms).
\newblock Master's thesis, Rosenheim Technical University of Applied Sciences.
\newblock Advised and supervised by Prof. Dr. Gerd Beneken and Prof. Dr. Marcel
  Tilly.

\bibitem[{Chi et~al.(2023)Chi, Chen, Chang, Lee, and
  Chang}]{chi-different-complexity}
Alison Chi, Li-Kuang Chen, Yi-Chen Chang, Shu-Hui Lee, and Jason~S. Chang.
  2023.
\newblock \href {https://doi.org/10.1162/tacl_a_00606} {Learning to paraphrase
  sentences to different complexity levels}.
\newblock \emph{Transactions of the Association for Computational Linguistics},
  11:1332--1354.

\bibitem[{Chung et~al.(2024)Chung, Hou, Longpre, Zoph, Tay, Fedus, Li, Wang,
  Dehghani, Brahma, Webson, Gu, Dai, Suzgun, Chen, Chowdhery, Castro-Ros,
  Pellat, Robinson, Valter, Narang, Mishra, Yu, Zhao, Huang, Dai, Yu, Petrov,
  Chi, Dean, Devlin, Roberts, Zhou, Le, and Wei}]{chung-flan-t5}
Hyung~Won Chung, Le~Hou, Shayne Longpre, Barret Zoph, Yi~Tay, William Fedus,
  Yunxuan Li, Xuezhi Wang, Mostafa Dehghani, Siddhartha Brahma, Albert Webson,
  Shixiang~Shane Gu, Zhuyun Dai, Mirac Suzgun, Xinyun Chen, Aakanksha
  Chowdhery, Alex Castro-Ros, Marie Pellat, Kevin Robinson, Dasha Valter,
  Sharan Narang, Gaurav Mishra, Adams Yu, Vincent Zhao, Yanping Huang, Andrew
  Dai, Hongkun Yu, Slav Petrov, Ed~H. Chi, Jeff Dean, Jacob Devlin, Adam
  Roberts, Denny Zhou, Quoc~V. Le, and Jason Wei. 2024.
\newblock \href {http://jmlr.org/papers/v25/23-0870.html} {Scaling
  instruction-finetuned language models}.
\newblock \emph{Journal of Machine Learning Research}, 25(70):1--53.

\bibitem[{Council~of Europe(2001)}]{council-cefr}
Council for Cultural Co-operation. Education Committee. Modern
  Languages~Division Council~of Europe. 2001.
\newblock \emph{Common European framework of reference for languages: Learning,
  teaching, assessment}.
\newblock Cambridge University Press.

\bibitem[{{DIN-Normenausschuss Ergonomie}(2024)}]{din-es}
{DIN-Normenausschuss Ergonomie}. 2024.
\newblock \href {https://doi.org/https://dx.doi.org/10.31030/3523268} {Einfache
  {Sprache} - {Anwendung} für das {Deutsche} - {Teil} 1: Sprachspezifische
  {Festlegungen} (plain language - application for the german language - part
  1: Language-specific provisions, {DIN} 8581-1)}.

\bibitem[{{DIN-Normenausschuss Ergonomie}(2025)}]{din-spec-ls}
{DIN-Normenausschuss Ergonomie}. 2025.
\newblock \href {https://doi.org/https://dx.doi.org/10.31030/3594547}
  {Empfehlungen für {Deutsche} {Leichte} {Sprache} (guidance for {German}
  {Easy} {Language}, {DIN} {SPEC} 33429)}.

\bibitem[{Engelmann et~al.(2024)Engelmann, Kreutz, Haak, and
  Schaer}]{engelmann-arts}
Bj{\"o}rn Engelmann, Christin~Katharina Kreutz, Fabian Haak, and Philipp
  Schaer. 2024.
\newblock \href {https://doi.org/10.18653/v1/2024.findings-emnlp.877} {{ARTS}:
  Assessing readability {\&} text simplicity}.
\newblock In \emph{Findings of the Association for Computational Linguistics:
  EMNLP 2024}, pages 14925--14942, Miami, Florida, USA. Association for
  Computational Linguistics.

\bibitem[{Farajidizaji et~al.(2024)Farajidizaji, Raina, and
  Gales}]{farajidizaji-readability-control}
Asma Farajidizaji, Vatsal Raina, and Mark Gales. 2024.
\newblock \href {https://aclanthology.org/2024.lrec-main.815/} {Is it possible
  to modify text to a target readability level? an initial investigation using
  zero-shot large language models}.
\newblock In \emph{Proceedings of the 2024 Joint International Conference on
  Computational Linguistics, Language Resources and Evaluation (LREC-COLING
  2024)}, pages 9325--9339, Torino, Italia. ELRA and ICCL.

\bibitem[{Fruth et~al.(2024)Fruth, Jegan, and Henrich}]{fruth-unsupervised-RF}
Leon Fruth, Robin Jegan, and Andreas Henrich. 2024.
\newblock \href {https://aclanthology.org/2024.determit-1.8/} {An approach
  towards unsupervised text simplification on paragraph-level for {G}erman
  texts}.
\newblock In \emph{Proceedings of the Workshop on DeTermIt! Evaluating Text
  Difficulty in a Multilingual Context @ LREC-COLING 2024}, pages 77--89,
  Torino, Italia. ELRA and ICCL.

\bibitem[{Gao et~al.(2025)Gao, Johnson, Froehlich, Carrer, and
  Ebling}]{gao-target-eval}
Yingqiang Gao, Kaede Johnson, David Froehlich, Luisa Carrer, and Sarah Ebling.
  2025.
\newblock \href {https://arxiv.org/abs/2507.01479} {Evaluating the
  effectiveness of direct preference optimization for personalizing german
  automatic text simplifications for persons with intellectual disabilities}.
\newblock \emph{Preprint}, arXiv:2507.01479.

\bibitem[{Grattafiori et~al.(2024)Grattafiori, Dubey, Jauhri, Pandey, Kadian,
  Al-Dahle, Letman, Mathur, Schelten, Yang, Fan, Goyal, Hartshorn, Yang, Mitra,
  Sravankumar, Korenev, Hinsvark, Rao, Zhang, Rodriguez, Gregerson, and
  et~al.}]{dubey-llama3}
Aaron Grattafiori, Abhimanyu Dubey, Abhinav Jauhri, Abhinav Pandey, Abhishek
  Kadian, Ahmad Al-Dahle, Aiesha Letman, Akhil Mathur, Alan Schelten, Amy Yang,
  Angela Fan, Anirudh Goyal, Anthony Hartshorn, Aobo Yang, Archi Mitra, Archie
  Sravankumar, Artem Korenev, Arthur Hinsvark, Arun Rao, Aston Zhang, Aurelien
  Rodriguez, Austen Gregerson, and Ava~Spataru et~al. 2024.
\newblock \href {https://arxiv.org/abs/2407.21783} {The llama 3 herd of
  models}.
\newblock \emph{Preprint}, arXiv:2407.21783.

\bibitem[{Heine(2017)}]{heine-gaf-ls}
Antje Heine. 2017.
\newblock Deutsch als {F}remd- und {Z}weitsprache – eine besondere {F}orm
  {L}eichter {S}prache? Überlegungen aus der {P}erspektive des {F}aches
  {DaF}/{DaZ}”.
\newblock \emph{„Leichte Sprache “im Spiegel theoretischer und angewandter
  Forschung}, pages 401--414.

\bibitem[{Hewett et~al.(2024)Hewett, Asghari, and Stede}]{hewett-elaborative}
Freya Hewett, Hadi Asghari, and Manfred Stede. 2024.
\newblock \href {https://doi.org/10.18653/v1/2024.sigdial-1.3} {Elaborative
  simplification for {G}erman-language texts}.
\newblock In \emph{Proceedings of the 25th Annual Meeting of the Special
  Interest Group on Discourse and Dialogue}, pages 29--39, Kyoto, Japan.
  Association for Computational Linguistics.

\bibitem[{Honnibal et~al.(2020)Honnibal, Montani, Landeghem, and
  Boyd}]{honnibal-spacy}
Matthew Honnibal, Ines Montani, Sofie~Van Landeghem, and Adriane Boyd. 2020.
\newblock \href {https://doi.org/10.5281/zenodo.1212303} {spacy:
  Industrial-strength natural language processing in python}.

\bibitem[{Hu et~al.(2022)Hu, yelong shen, Wallis, Allen-Zhu, Li, Wang, Wang,
  and Chen}]{hulora}
Edward~J Hu, yelong shen, Phillip Wallis, Zeyuan Allen-Zhu, Yuanzhi Li, Shean
  Wang, Lu~Wang, and Weizhu Chen. 2022.
\newblock \href {https://openreview.net/forum?id=nZeVKeeFYf9} {Lo{RA}: Low-rank
  adaptation of large language models}.
\newblock In \emph{International Conference on Learning Representations}.

\bibitem[{Kl{\"o}ser et~al.(2024)Kl{\"o}ser, Beele, Schagen, and
  Kraft}]{kloser-synthetic}
Lars Kl{\"o}ser, Mika Beele, Jan-Niklas Schagen, and Bodo Kraft. 2024.
\newblock \href {https://aclanthology.org/2024.ltedi-1.7/} {{G}erman text
  simplification: Finetuning large language models with semi-synthetic data}.
\newblock In \emph{Proceedings of the Fourth Workshop on Language Technology
  for Equality, Diversity, Inclusion}, pages 63--72, St. Julian's, Malta.
  Association for Computational Linguistics.

\bibitem[{Krippendorff(2011)}]{krippendorff-alpha}
Klaus Krippendorff. 2011.
\newblock Computing krippendorff’s alpha-reliability.

\bibitem[{Liu et~al.(2025)Liu, Nam, Cui, and
  Swayamdipta}]{liu2025-llm-judge-simplification}
Joseph Liu, Yoonsoo Nam, Xinyue Cui, and Swabha Swayamdipta. 2025.
\newblock \href {https://arxiv.org/abs/2504.09394} {Evaluation under imperfect
  benchmarks and ratings: A case study in text simplification}.
\newblock \emph{Preprint}, arXiv:2504.09394.

\bibitem[{Maa{\ss}(2020)}]{maass-easy-plain-plus}
Christiane Maa{\ss}. 2020.
\newblock \href {https://doi.org/10.26530/20.500.12657/42089} {\emph{Easy
  language--plain language--easy language plus: Balancing comprehensibility and
  acceptability}}.
\newblock Frank \& Timme, Berlin.

\bibitem[{Madina et~al.(2023)Madina, Gonzalez-Dios, and
  Siegel}]{madina-lectura-facil}
Margot Madina, Itziar Gonzalez-Dios, and Melanie Siegel. 2023.
\newblock \href {https://doi.org/10.1145/3594806.3596530} {Easy-to-read
  language resources and tools for three european languages}.
\newblock In \emph{Proceedings of the 16th International Conference on
  PErvasive Technologies Related to Assistive Environments}, PETRA '23, page
  693–699, New York, NY, USA. Association for Computing Machinery.

\bibitem[{Malik et~al.(2024)Malik, Mayhew, Piech, and
  Bicknell}]{malik-gpt4-synthetic}
Ali Malik, Stephen Mayhew, Christopher Piech, and Klinton Bicknell. 2024.
\newblock \href {https://doi.org/10.18653/v1/2024.findings-acl.926} {From
  tarzan to {T}olkien: Controlling the language proficiency level of {LLM}s for
  content generation}.
\newblock In \emph{Findings of the Association for Computational Linguistics:
  ACL 2024}, pages 15670--15693, Bangkok, Thailand. Association for
  Computational Linguistics.

\bibitem[{Mallinson et~al.(2020)Mallinson, Sennrich, and
  Lapata}]{Mallinson-geolino}
Jonathan Mallinson, Rico Sennrich, and Mirella Lapata. 2020.
\newblock \href {https://doi.org/10.5167/uzh-191668} {Zero-shot crosslingual
  sentence simplification}.
\newblock In \emph{Proceedings of the 2020 Conference on Empirical Methods in
  Natural Language Processing (EMNLP)}, pages 5109--5126, Online. Association
  for Computational Linguistics.

\bibitem[{Manning(2024)}]{manning-ki-tools}
Sabine Manning. 2024.
\newblock \href
  {https://multisprech.org/2024/05/22/ki-tools-fur-einfache-sprache-leistungen-von-10-tools-auf-einen-blick/}
  {Ki-tools für einfache sprache: Leistungen von 10 tools auf einen blick}.
\newblock Published 22.05.2024, last accessed 30.04.2025.

\bibitem[{Naderi et~al.(2019)Naderi, Mohtaj, Ensikat, and
  Möller}]{Naderi-dataset}
Babak Naderi, Salar Mohtaj, Kaspar Ensikat, and Sebastian Möller. 2019.
\newblock \href {https://doi.org/10.48550/ARXIV.1904.07733} {Subjective
  assessment of text complexity: A dataset for german language}.
\newblock \emph{arXiv preprint}.

\bibitem[{OECD(2013)}]{desjardins-oecd}
OECD. 2013.
\newblock Oecd skills outlook 2013: First results from the survey of adult
  skills.
\newblock OECD Publishing, \url{http://dx.doi.org/10.1787/9789264204256-en }.

\bibitem[{Papineni et~al.(2002)Papineni, Roukos, Ward, and Zhu}]{papineni-bleu}
Kishore Papineni, Salim Roukos, Todd Ward, and Wei-Jing Zhu. 2002.
\newblock \href {https://doi.org/10.3115/1073083.1073135} {{B}leu: a method for
  automatic evaluation of machine translation}.
\newblock In \emph{Proceedings of the 40th Annual Meeting of the Association
  for Computational Linguistics}, pages 311--318, Philadelphia, Pennsylvania,
  USA. Association for Computational Linguistics.

\bibitem[{Pfister et~al.(2025)Pfister, Wunderle, and Hotho}]{pfister-lammlein}
Jan Pfister, Julia Wunderle, and Andreas Hotho. 2025.
\newblock \href {https://doi.org/10.18653/v1/2025.acl-long.111}
  {{LL}{\"a}{M}mlein: Transparent, compact and competitive {G}erman-only
  language models from scratch}.
\newblock In \emph{Proceedings of the 63rd Annual Meeting of the Association
  for Computational Linguistics (Volume 1: Long Papers)}, pages 2227--2246,
  Vienna, Austria. Association for Computational Linguistics.

\bibitem[{Plüster(2023)}]{pluster-leoLM}
Björn Plüster. 2023.
\newblock \href {https://laion.ai/blog/leo-lm/} {Leolm: Igniting
  german-language llm research}.

\bibitem[{S\"{a}uberli et~al.(2024)S\"{a}uberli, Holzknecht, Haller, Deilen,
  Schiffl, Hansen-Schirra, and Ebling}]{sauberli-eye-tracking}
Andreas S\"{a}uberli, Franz Holzknecht, Patrick Haller, Silvana Deilen, Laura
  Schiffl, Silvia Hansen-Schirra, and Sarah Ebling. 2024.
\newblock \href {https://doi.org/10.1145/3613904.3642570} {Digital
  comprehensibility assessment of simplified texts among persons with
  intellectual disabilities}.
\newblock In \emph{Proceedings of the 2024 CHI Conference on Human Factors in
  Computing Systems}, CHI '24, New York, NY, USA. Association for Computing
  Machinery.

\bibitem[{Schomacker et~al.(2023)Schomacker, Gille, Tropmann-Frick, and von~der
  H{\"u}lls}]{schomacker-data}
Thorben Schomacker, Michael Gille, Marina Tropmann-Frick, and J{\"o}rg von~der
  H{\"u}lls. 2023.
\newblock \href {https://aclanthology.org/2023.konvens-main.6/} {Data and
  approaches for {G}erman text simplification {--} towards an
  accessibility-enhanced communication}.
\newblock In \emph{Proceedings of the 19th Conference on Natural Language
  Processing (KONVENS 2023)}, pages 63--68, Ingolstadt, Germany. Association
  for Computational Lingustics.

\bibitem[{Spring et~al.(2021)Spring, Rios, and
  Ebling}]{spring-multi-level-simp}
Nicolas Spring, Annette Rios, and Sarah Ebling. 2021.
\newblock \href {https://aclanthology.org/2021.ranlp-1.150/} {Exploring
  {G}erman multi-level text simplification}.
\newblock In \emph{Proceedings of the International Conference on Recent
  Advances in Natural Language Processing (RANLP 2021)}, pages 1339--1349, Held
  Online. INCOMA Ltd.

\bibitem[{Stajner(2021)}]{stajner-social-good}
Sanja Stajner. 2021.
\newblock \href {https://doi.org/10.18653/v1/2021.findings-acl.233} {Automatic
  text simplification for social good: Progress and challenges}.
\newblock In \emph{Findings of the Association for Computational Linguistics:
  ACL-IJCNLP 2021}, pages 2637--2652, Online. Association for Computational
  Linguistics.

\bibitem[{Stodden(2024{\natexlab{a}})}]{stodden-easse-de}
Regina Stodden. 2024{\natexlab{a}}.
\newblock \href {https://doi.org/10.18653/v1/2024.tsar-1.11} {{EASSE}-{DE} {\&}
  {EASSE}-multi: Easier automatic sentence simplification evaluation for
  {G}erman {\&} multiple languages}.
\newblock In \emph{Proceedings of the Third Workshop on Text Simplification,
  Accessibility and Readability (TSAR 2024)}, pages 107--116, Miami, Florida,
  USA. Association for Computational Linguistics.

\bibitem[{Stodden(2024{\natexlab{b}})}]{stodden-mt5-deplain}
Regina Stodden. 2024{\natexlab{b}}.
\newblock \href {https://aclanthology.org/2024.determit-1.1/} {Reproduction
  {\&} benchmarking of {G}erman text simplification systems}.
\newblock In \emph{Proceedings of the Workshop on DeTermIt! Evaluating Text
  Difficulty in a Multilingual Context @ LREC-COLING 2024}, pages 1--15,
  Torino, Italia. ELRA and ICCL.

\bibitem[{Stodden et~al.(2023)Stodden, Momen, and Kallmeyer}]{stodden-deplain}
Regina Stodden, Omar Momen, and Laura Kallmeyer. 2023.
\newblock \href {https://doi.org/10.18653/v1/2023.acl-long.908} {{DE}plain: A
  {G}erman parallel corpus with intralingual translations into plain language
  for sentence and document simplification}.
\newblock In \emph{Proceedings of the 61st Annual Meeting of the Association
  for Computational Linguistics (Volume 1: Long Papers)}, pages 16441--16463,
  Toronto, Canada. Association for Computational Linguistics.

\bibitem[{{Thanh Mai Pham}(2024)}]{Pham-MA-German4All}
{Thanh Mai Pham}. 2024.
\newblock German4all: Paraphrasing german texts to different complexity levels
  with gpt-generated synthetic data.
\newblock Master's thesis, {Technical University of Munich}.
\newblock Advised and supervised by Miriam Ansch{\"u}tz and Georg Groh.

\bibitem[{Toborek et~al.(2023)Toborek, Busch, Bo{\ss}ert, Bauckhage, and
  Welke}]{toborek-simple-corpus}
Vanessa Toborek, Moritz Busch, Malte Bo{\ss}ert, Christian Bauckhage, and
  Pascal Welke. 2023.
\newblock \href {https://doi.org/10.18653/v1/2023.acl-long.638} {A new aligned
  simple {G}erman corpus}.
\newblock In \emph{Proceedings of the 61st Annual Meeting of the Association
  for Computational Linguistics (Volume 1: Long Papers)}, pages 11393--11412,
  Toronto, Canada. Association for Computational Linguistics.

\bibitem[{Toshevska and Gievska(2025)}]{Toshevska-finetuning-prompting}
Martina Toshevska and Sonja Gievska. 2025.
\newblock \href {https://doi.org/10.1109/ACCESS.2025.3548967} {Llm-based text
  style transfer: Have we taken a step forward?}
\newblock \emph{IEEE Access}, 13:44707--44721.

\bibitem[{Trienes et~al.(2024)Trienes, Joseph, Schl{\"o}tterer, Seifert, Lo,
  Xu, Wallace, and Li}]{trienes-infolossQA}
Jan Trienes, Sebastian Joseph, J{\"o}rg Schl{\"o}tterer, Christin Seifert, Kyle
  Lo, Wei Xu, Byron~C. Wallace, and Junyi~Jessy Li. 2024.
\newblock \href {https://aclanthology.org/2024.acl-long.234} {{InfoLossQA}:
  {C}haracterizing and recovering information loss in text simplification}.
\newblock In \emph{Proceedings of the 62nd Annual Meeting of the Association
  for Computational Linguistics}, pages 4263--4294.

\bibitem[{Xu et~al.(2016)Xu, Napoles, Pavlick, Chen, and
  Callison-Burch}]{Xu-SARI}
Wei Xu, Courtney Napoles, Ellie Pavlick, Quanze Chen, and Chris Callison-Burch.
  2016.
\newblock \href {https://doi.org/10.1162/tacl_a_00107} {Optimizing statistical
  machine translation for text simplification}.
\newblock \emph{Transactions of the Association for Computational Linguistics},
  4:401--415.

\bibitem[{Zhang* et~al.(2020)Zhang*, Kishore*, Wu*, Weinberger, and
  Artzi}]{Zhang-BERTScore}
Tianyi Zhang*, Varsha Kishore*, Felix Wu*, Kilian~Q. Weinberger, and Yoav
  Artzi. 2020.
\newblock \href {https://openreview.net/forum?id=SkeHuCVFDr} {Bertscore:
  Evaluating text generation with bert}.
\newblock In \emph{International Conference on Learning Representations}.

\end{thebibliography}

\clearpage
\appendix

\section{Complexity level specifications}\label{sec:CL_levels}
\begin{table*}
    \centering
    \begin{tabularx}{\linewidth}{XXlXXX}\toprule
        \textbf{Human operation} & \textbf{LLM Judge criterion} & \textbf{\href{https://huggingface.co/microsoft/phi-4}{Phi-4}} & \textbf{\href{https://huggingface.co/meta-llama/Llama-3.3-70B-Instruct}{Llama-3.3-70B-Instruct}} & \textbf{\href{Qwen/Qwen2.5- 72B-Instruct}{Qwen2.5-72B-Instruct}} %& \textbf{R1-Qwen} 
        & \textbf{\href{https://huggingface.co/google/gemma-3-27b-it}{Gemma-3-27B-it}}\\
        \midrule
        \textbf{corrected info} & \textbf{content preservation}  & \textbf{-0.23 (0.007)}  & -0.05 (0.566)  & \textbf{-0.24 (0.006)}  & \textbf{-0.27 (0.001)} \\
\textbf{removed info} & \textbf{information addition}  & \textbf{0.38 (0.000)}  & 0.13 (0.126)  & \textbf{0.30 (0.001)}  & \textbf{0.51 (0.000)} \\
\textbf{hallucination in origin} & \textbf{information addition}  & \textbf{0.23 (0.008)}  & 0.10 (0.231)  & 0.20 (0.022)  & \textbf{0.35 (0.000)} \\
\textbf{added info} & \textbf{information loss}  & 0.14 (0.101)  & 0.13 (0.130)  & 0.16 (0.059)  & 0.13 (0.153) \\
\textbf{adjusted complexity} & \textbf{complexity level}  & -0.07 (0.401)  & 0.19 (0.031)  & \textbf{0.32 (0.000)}  & \textbf{0.34 (0.000)} \\
    \bottomrule
    \end{tabularx}
    \caption{Spearman correlation strength and p-values (in brackets) of human operations in the \textit{German4All-annotated} dataset and the LLM judge criteria. Statistically significant correlations (p-values $\leq0.01$) are bolded.} 
    \label{tab:judge_ablation}
\end{table*}

We distinguish between these five complexity levels. The following descriptions were originally provided in German, but translated to English for the paper.
\begin{enumerate}
    \item Leichte Sprache Plus (literal translation: Easy Language Plus) \begin{itemize}[leftmargin=*]
        \item \textit{Target group}: People with reading difficulties, including people with learning disabilities and those who have only recently started to learn German.
        \item \textit{Characteristics}: Very short sentences, only short and frequently used words, direct speech, avoidance of abbreviations, metaphors, or irony.
        \item \textit{Examples areas}: simple instructions, accessible websites.
    \end{itemize}
    \item Simple German for beginners 
    \begin{itemize}[leftmargin=*]
        \item \textit{Target group}: Non-native speakers with basic knowledge of German.
        \item \textit{Characteristics}: Simple sentence structures, basic vocabulary, strong focus on important information, avoidance of culture-specific expressions.
        \item \textit{Example areas}: Language learning materials, introductory web texts.
    \end{itemize}
    \item Commonly used language
    \begin{itemize}[leftmargin=*]
        \item \textit{Target group}: General public with different levels of education.
        \item \textit{Characteristics}: Clear, structured sentences, focus on comprehensibility, avoidance of technical terms.
        \item \textit{Example areas}: Wide-ranging news portals, blogs.
    \end{itemize}
    \item Elevated everyday language
    \begin{itemize}[leftmargin=*]
        \item \textit{Target group}: Regular readers with a good understanding of the language.
        \item \textit{Characteristics}: More varied vocabulary, occasional technical terminology with explanations, complex sentence structures.
        \item \textit{Example areas}: Specialist blogs, quality newspapers.
    \end{itemize}
    \item Academic language
    \begin{itemize}[leftmargin=*]
        \item \textit{Target group}: Academics and experts.
        \item \textit{Characteristics}: Complex sentence structures, specialized terminology, use of technical terms.
        \item \textit{Example areas}: Specialist journals, scientific publications.
    \end{itemize}
\end{enumerate}

\section{LLM judge ablation}\label{app:llm_ablation_models}
We compared multiple LLMs to see if they properly capture our evaluation criteria. For this, we selected four recent multilingual open-source models. The annotations of human operations in the \textit{German4All-annotated} dataset give an indication about the errors in the GPT-4 outputs. For example, if the human manually removed information from the output, that means that GPT-4 has added information that should not be there. Similarly, if the human has adjusted the text complexity, that means the previous version of the text was improper for the target complexity level. We selected five human operations in the dataset and paired them with criteria in our evaluation setup. An ideal judge model should have high correlations for all these pairs. The results are presented in \autoref{tab:judge_ablation}.

Our prompt features a sanity check on whether the LLM understood the prompt. As such, the answer to question 4 (type of addition) should be \textit{'NaN'} if the answer to question 3 was \textit{never}. All models succeeded with this prompt understanding test. Concerning the human annotations, no LLM judge can properly detect information that was missing from the GPT-4 outputs. However, Phi4 and Gemma3 can detect all other content-related errors, like hallucinations and incorrect information. For the complexity evaluation, only Qwen2.5 and Gemma3 have a statistically significant correlation with the human annotations. Therefore, Gemma3 is our preferred judge model and was used for testing the full dataset.

\section{Distillation training parameters}\label{sec:params}
For model fine-tuning, we used \href{https://pypi.org/project/transformers/}{transformers} 4.52.4 and \href{https://pypi.org/project/peft/}{peft} 0.15.2 with the following configuration:
\begin{lstlisting}[language=python]
lora_config = LoraConfig(
    r=32,
    lora_alpha=16,
    target_modules=["q", "v", "k", "o"],
    lora_dropout=0.05,
    bias="none",
    task_type=TaskType.SEQ_2_SEQ_LM,
)

tr_args = Seq2SeqTrainingArguments(
    output_dir=output_dir,
    eval_strategy="steps",
    save_strategy="steps",
    save_steps=20000,
    eval_steps=100000,
    learning_rate=3e-4,
    per_device_train_batch_size=64,
    per_device_eval_batch_size=64,
    auto_find_batch_size=True,
    weight_decay=0.01,
    num_train_epochs=2,
    predict_with_generate=True,
    logging_dir="./logs",
    logging_steps=1000,
    bf16=True,
    fp16=False,
    report_to="none",
    lr_scheduler_type="linear", 
    warmup_steps=2750
)
\end{lstlisting}
\section{Benchmarking German ATS models on our dataset}\label{app:g4a_benchmark}
We compare our model against the best-performing models according to \citet{stodden-mt5-deplain} in terms of simplicity and SARI scores, as well as two more recent German simplification models. All models were fine-tuned on different datasets, and the base versions range from mBart to Llama3 \citep{dubey-llama3} and the German-specific Leo-LM \citep{pluster-leoLM}.

The highest readability score is obtained by our German4All model, prompted to output texts at level 1, while the erlesen model shows the lowest. Mt5-SGC has the lowest compression rate, mostly because it just outputs one sentence, no matter the input length. Our model learns the nuances between simplifications to levels 1 and 2 and achieves higher BLEU, SARI, and BERTscore, i.e., is closer to the reference, when creating outputs of the respective level.
\begin{table*}[ht]
    \centering
    \small
    \begin{tabular}{lccccccc}\toprule
         \textbf{Model} & \textbf{BLEU}$\uparrow$ & \textbf{SARI}$\uparrow$ &  \textbf{BS\_F1}$\uparrow$ & \textbf{FRE}$\uparrow$ & \textbf{Compression}$\downarrow$ & \textbf{Sent. splits}$\uparrow$ & \textbf{Copies}$\downarrow$ \\ \midrule
         \multicolumn{8}{c}{\textit{German4All - Target complexity level 1}} \\
        References & - & - & - & 75.7 & 0.73 & 1.96 & 0.0\\\hdashline \\[-1.7ex]
        erlesen-leo-7b & 4.55 & 38.15 & 0.22 & 49.25 & 1.08 & 1.1 & 0.01 \\
        mBART-DEplain-APA+web & 2.74 & 35.88 & 0.26 & 52.1 & 0.49 & 0.72 & 0.0\\
        mt5-SGC & 0.2 & 38.65 & 0.2 & 70.5 & \textbf{0.22} & 0.51 & 0.0\\
        Simba & 4.85 & 41.29 & 0.28 & 65.09 & 0.71 & 1.48 & 0.0 \\
        German4All-level1 (ours) & \textbf{14.47} & \textbf{53.9} & \textbf{0.45} & \textbf{77.2} & 0.69 & \textbf{1.99} & 0.0\\
        German4All-level2 (ours) & 10.85 & 47.43 & 0.41 & 67.18 & 0.8 & 1.49 & 0.0\\ \midrule
         \multicolumn{8}{c}{\textit{German4All - Target complexity level 2}} \\
        References & - & - & - & 64.02 & 0.82 & 1.45 & 0.0 \\\hdashline \\[-1.7ex]
        erlesen-leo-7b & 8.44 & 42.42 & 0.28 & 49.25 & 1.08 & 1.1 & 0.01\\
        mBART-DEplain-APA+web & 4.22 & 37.98 & 0.35 & 52.1 & 0.49 & 0.72 & 0.0\\
        mt5-SGC &0.19 & 36.86 & 0.2 & 70.5 & \textbf{0.22} & 0.51 & 0.0 \\
        Simba & 5.59 & 41.93 & 0.33 & 65.09 & 0.71 & 1.48 & 0.0 \\
        German4All-level1 (ours) & 10.85 & 50.41 & 0.41 & \textbf{77.2} & 0.69 & \textbf{1.99} & 0.0 \\
        German4All-level2 (ours) & \textbf{15.66} & \textbf{51.7} & \textbf{0.48} & 67.18 & 0.8 & 1.49 & 0.0 \\
        \bottomrule
    \end{tabular}
    \caption{Performance comparison of different German ATS models on our German4All-Corrected test set. Best values are bolded. Our models show the highest scores and best fit the characteristics of the references in the respective complexity level.}
    \label{tab:German4All_benchmark}
\end{table*}

\section{Further example}\label{app:judge_example}
Here is an example from the main corpus and the respective LLM judge annotations: 
\paragraph{Orig text:} "Die vier zum Teil beschaädigten Nischenfiguren der Südfassade unterhalb des Giebels stellen Drama, Gesang, Musik und Tanz dar. Sie wurden von dem Stuttgarter Bildhauer Theodor Bausch geschaffen." ({\small \textit{EN: The four partially damaged niche figures on the south facade below the gable represent drama, singing, music, and dance. They were created by the Stuttgart sculptor Theodor Bausch.}})

\textbf{Level 1}: "An einem großen Haus gibt es Figuren aus Stein. Sie sind ein bisschen kaputt. Die Figuren zeigen vier Dinge: ein Theaterstück, Singen, Musik und Tanzen. Ein Mann namens Theodor Bausch hat sie gemacht." ({\small \textit{EN: There are stone figures on a large house. They are a little damaged. The figures depict four things: a play, singing, music, and dancing. A man named Theodor Bausch made them.}})\\
$\rightarrow$ LLM judge annotations: content\_preservation = approximately, information\_loss = often, information\_addition = seldom,  complexity\_level = a bit too complicated

\textbf{Level 2}: An der Südseite eines Gebaudes gibt es vier beschadigte Statuen. Diese Statuen zeigen vier Künste: Schauspiel, Gesang, Musik und Tanz. Ein Künstler aus Stuttgart, Theodor Bausch, hat sie erschaffen. ({\small \textit{There are four damaged statues on the south side of a building. These statues depict four arts: acting, singing, music, and dance. They were created by an artist from Stuttgart named Theodor Bausch.}})\\
$\rightarrow$ LLM judge annotations: content\_preservation = approximately, information\_loss = sometimes, information\_addition = seldom,  complexity\_level = a bit too easy

\textbf{Level 3}: An der Südwand eines Gebaudes befinden sich vier teilweise beschadigte Statuen. Sie symbolisieren Dramatik, Gesang, Musik und Tanz. Diese Kunstwerke wurden vom Stuttgarter Bildhauer Theodor Bausch geschaffen. ({\small \textit{EN: On the south wall of a building there are four partially damaged statues. They symbolize drama, singing, music, and dance. These works of art were created by the Stuttgart sculptor Theodor Bausch.}})\\
$\rightarrow$ LLM judge annotations: content\_preservation = correct, information\_loss = seldom, information\_addition = seldom,  complexity\_level = appropriate

\textbf{Level 4}: Die Südfassade des Gebaudes ziert eine Reihe von Nischenfiguren, die trotz teils sichtbarer Schaden die Künste des Dramas, Gesanges, der Musik und des Tanzes darstellen. Sie sind Werke des renommierten Stuttgarter Bildhauers Theodor Bausch. ({\small \textit{EN: The south facade of the building is adorned with a series of niche figures which, despite some visible damage, represent the arts of drama, singing, music, and dance. They are the work of the renowned Stuttgart sculptor Theodor Bausch.}})\\
$\rightarrow$ LLM judge annotations: content\_preservation = correct, information\_loss = never, information\_addition = seldom,  complexity\_level = appropriate

\textbf{Level 5}: Die Südfassade des architektonischen Werkes weist vier beschadigte Nischenfiguren auf, welche die Disziplinen Drama, Gesang, Musik und Tanz reprasentieren. Diese künstlerischen Darstellungen wurden von Theodor Bausch, einem Bildhauer aus Stuttgart, gefertigt und reflektieren kulturelle Ausdrucksformen der bildenden Künste. ({\small \textit{EN: The south facade of the architectural work features four damaged niche figures representing the disciplines of drama, singing, music, and dance. These artistic representations were created by Theodor Bausch, a sculptor from Stuttgart, and reflect cultural forms of expression in the fine arts.}})\\
$\rightarrow$ LLM judge annotations: content\_preservation = approximately, information\_loss = sometimes, information\_addition = often,  complexity\_level = a bit too complicated

The samples are all of high quality. However, levels 3 and 4 achieve the best content preservation, while level 1 removes information ("Südfassade unterhalb des Giebels" ({\small \textit{south facade below the gable}}), "Stuttgarter") and level 5 adds information ("architektonischen Werkes" ({\small \textit{architectural work}}), "reflektieren kulturelle Ausdrucksformen der bildenden Künste" ({\small \textit{reflect cultural forms of expression in the fine arts}})).

\section{System prompts}
The following figures show the prompts that we used to generate the dataset and evaluate it using an LLM judge. All prompts were provided in German to avoid code switching, but were translated to English for the paper.
\subsection{Synthetic data generation}\label{app:synt_data_prompt}
% \begin{figure}[ht]
% \centering
% \begin{adjustbox}{width=\textwidth}
% \begin{tcolorbox}[paragraphbox]
%     \begin{lstlisting}[basicstyle =\footnotesize\ttfamily]
% You are an expert in adapting texts to different complexity levels.
%     \end{lstlisting}
% \end{tcolorbox}
% \end{adjustbox}
% \caption{English translation of the system message for the paraphrase generation.}
% \label{fig:system-prompt}
% \end{figure}

\begin{figure*}[hb]
\centering
%\begin{adjustbox}{width=0.95\linewidth}
\begin{tcolorbox}[backgroundbox]
        \begin{tcolorbox}[paragraphbox]
        % \noindent
        \begin{lstlisting}
**Context**
There are five Complexity levels: <definitions from App. A>
        \end{lstlisting}
        \end{tcolorbox}

        \begin{tcolorbox}[paragraphbox]
        \begin{lstlisting}[]
**Example**
Text: <input text>
Paraphrases in json Format:
{
    "1": "Paraphrase at complexity level 1",
    "2": "Paraphrase at complexity level 2",
    "3": "Paraphrase at complexity level 3",
    "4": "Paraphrase at complexity level 4",
    "5": "Paraphrase at complexity level 5"
}
        \end{lstlisting}
        \end{tcolorbox}

        \begin{tcolorbox}[paragraphbox]
        \begin{lstlisting}
**Task**
Paraphrase the given text to five different complexity levels. When creating the paraphrases, you should proceed step by step,considering the target group and the specific characteristics and areas of application of each complexity level. Do this task in the form of an inner monologue. Do not explain your thought process, but present the final paraphrased texts directly.
        \end{lstlisting}

        \end{tcolorbox}

        \begin{tcolorbox}[paragraphbox]
        \begin{lstlisting}
Text: <input text>
        \end{lstlisting}
        \end{tcolorbox}

        \begin{tcolorbox}[paragraphbox]
        \begin{lstlisting}
**Response in json format**
{
    "1": "Level 1 text",
    "2": "Level 2 text",
    "3": "Level 3 text",
    "4": "Level 4 text",
    "5": "Level 5 text"
}

        \end{lstlisting}
        \end{tcolorbox}
    \end{tcolorbox}

%\end{adjustbox}
\caption{Prompt structure for generating paraphrases at five complexity levels. Colored boxes separate each section. First, we describe each complexity level. Then, we show a 1-shot example. Afterward, we provide the task description, the input text, and the response format.}
\label{fig:prompt-design}
\end{figure*}

\begin{figure*}[ht]
\centering
%\begin{adjustbox}{width=0.95\linewidth}
\begin{tcolorbox}[backgroundbox]
        \begin{tcolorbox}[paragraphbox]
        \begin{lstlisting}[]
**Beispiel**
Text: Die Ortschaft Danbury geht auf die Gründung einer vorchristlichen Wallburg zurück. Funde lassen auf eine erste Siedlung in der Eisenzeit schließen. Nach     Römern und Angelsachsen, wurde das Gebiet um Danbury im 11. Jahrhundert von       dänischen Stämmen erobert. Nach der Eroberung Englands durch Wilhelm den Eroberer 1066 wurde das Land um Danbury von den Normannen besiedelt. Das älteste, heute noch erhaltene Gebäude, ist die Kirche St. John the Baptist, die im 13. Jahrhundert errichtet wurde.
Paraphrasierungen im json Format:
{
    "1": "Danbury ist ein Ort. In Danbury gab es früher eine große alte Burg. Viele verschiedene Leute haben dort gelebt. Die ersten Menschen kamen vor sehr langer Zeit. Später kamen Menschen aus Dänemark und dann aus Frankreich. In Danbury steht eine sehr alte Kirche. Sie ist ungefähr 800 Jahre alt.",
    "2": "Danbury ist ein Ort mit einer alten Burg, die noch vor der christlichen Zeit gebaut wurde. Zuerst lebten dort Menschen in der Eisenzeit. Dann kamen   Römer, Angelsachsen und später Dänen. Nach einer großen Schlacht kamen Menschen aus Frankreich, die Normannen. Die älteste Kirche dort heißt St. John the Baptist und wurde im Mittelalter gebaut.", 
    "3": "Danbury ist bekannt für seine historische Burg, die vor der christlichen Ära errichtet wurde. Die ersten Bewohner kamen während der Eisenzeit. über die Jahrhunderte hinweg wurde das Gebiet von Römern, Angelsachsen und später von Dänen bewohnt. Nach der normannischen Eroberung Englands im Jahr 1066 wurde Danbury von den Normannen übernommen. Die Kirche St. John the Baptist, die älteste in Danbury, stammt aus dem 13. Jahrhundert.",
    "4": "Danbury besitzt eine lange Geschichte, die mit einer prähistorischen Festung beginnt. Archäologische Funde deuten darauf hin, dass die Region bereits in der Eisenzeit bewohnt war. Nach den Römern und Angelsachsen kamen im 11. Jahrhundert dänische Eroberer. Die normannische Eroberung Englands im Jahr 1066 brachte weitere Veränderungen mit sich, und Danbury fiel in die     Hände der Normannen. Die Kirche St. John the Baptist, erbaut im 13. Jahrhundert, ist das älteste noch bestehende Gebäude.",
    "5": "Die historische Entwicklung von Danbury lässt sich bis zu einer prächristlichen Festungsanlage zurückverfolgen. Archäologische Befunde belegen eine frühe Besiedlung während der Eisenzeit. Die Abfolge der Herrschaftswechsel von Römern zu Angelsachsen und später zu dänischen Stämmen im 11. Jahrhundert illustriert die komplexe Sozialstruktur dieser Ära. Mit der normannischen Eroberung Englands im Jahre 1066 wurde Danbury Teil eines erweiterten Herrschaftsgebietes. Die Kirche St. John the Baptist aus dem 13. Jahrhundert dient als architektonisches Zeugnis dieser vielschichtigen Geschichte."
}
        \end{lstlisting}
        \end{tcolorbox}
    \end{tcolorbox}

%\end{adjustbox}
\caption{German 1-shot example provided in the original prompt.}
\label{fig:german-1-shot}
\end{figure*}

\FloatBarrier
\enlargethispage{2\baselineskip}
\subsection{LLM judge}\label{app:llm_judge_prompt}
\begin{figure*}[!b]
\centering
%\begin{adjustbox}{width=0.95\linewidth}
\begin{tcolorbox}[backgroundbox]
        \begin{tcolorbox}[paragraphbox]
        % \noindent
        \begin{lstlisting}
You will receive an original text and a paraphrased version.

The paraphrases can be available in 5 different levels of difficulty. We define these levels as follows: <definitions from App. A>
        \end{lstlisting}
        \end{tcolorbox}

        \begin{tcolorbox}[paragraphbox]
        \begin{lstlisting}
Your task is to evaluate the paraphrasing under the following aspects:

- content_preservation: How well does the content of the paraphrase match the original text? Pay particular attention to whether the meaning has been changed or simplified so that nuances are lost or new interpretations are created.
- information_loss: Is information missing from the paraphrase that appears in the original?
- information_addition: Does the paraphrase contain additional information that does not appear in the original? This includes explanatory paraphrases or examples that introduce new elements of meaning.
- type_of_addition: If you answered 'never' to the previous question, answer 'NaN' here. Otherwise, if additional information is included, indicate what type it is: e.g., explanations, embellishments, or correct or incorrect facts.
- complexity_level: How well does the paraphrase match the given level of complexity? Pay attention to the linguistic features of the respective level, not just the content.

        \end{lstlisting}
        \end{tcolorbox}

        \begin{tcolorbox}[paragraphbox]
        \begin{lstlisting}
**IMPORTANT:**
- If abstract terms (e.g. 'universalist') are replaced by simple paraphrases (e.g. 'for all people'), this is considered an additional explanation.
- If the paraphrase creates a shift in content (e.g. from ideological concept to simple statement), this may constitute *factually incorrect information*.
- The assessment must be made exclusively in the following JSON format and must comply with this schema:
{json_schema}

Return **only** the JSON object -- without any further text.
        \end{lstlisting}

        \end{tcolorbox}
    \end{tcolorbox}

%\end{adjustbox}
\caption{LLM judge system prompt with the task description and additional guidance.}
\label{fig:prompt-design-judge}
\end{figure*}

\begin{figure*}[htbp]
\centering
\begin{tcolorbox}[backgroundbox]
        \begin{tcolorbox}[paragraphbox]
        \begin{lstlisting}
Evaluate this text:
Original text: <input text original text>
Paraphrased text: <input text paraphrased text>
Complexity level of the paraphrase: <input text complexity level>

Pay particular attention to whether the meaning is lost or changed by simplifications.
        \end{lstlisting}
        \end{tcolorbox}

        \begin{tcolorbox}[paragraphbox]
        \begin{lstlisting}
**Response in json format**
{
    'content_preservation': 'incorrectly'|'approximately'|'correctly'
    'information_loss': 'never'|'seldom'|'sometimes'|'often',
    'information_addition': 'never'|'seldom'|'sometimes'|'often',
    'type_of_addition': list[
        'embellishment'|'explanations/definitions'|
        'factually_incorrect_information'|'factually_correct_information'|
        'other'|'NaN'
    ],
  'complexity_level': 'too_easy'|'a_bit_too_easy'|'appropriate'|
  'a_bit_too_complicated'|'too_complicated',
}
        \end{lstlisting}
        \end{tcolorbox}
    \end{tcolorbox}

%\end{adjustbox}
\caption{LLM Judge user prompt and JSON output format.}
\label{fig:prompt-design-judge-user}
\end{figure*}
\end{document}